\documentclass[10pt,a4paper]{article}

\usepackage[numbers,authoryear,sort&compress]{natbib}

\usepackage{ragged2e,float}
\usepackage{graphicx,hyperref}
\usepackage{amsmath,amsthm,amssymb}
\usepackage{booktabs,multicol,multirow}
\usepackage[]{todonotes} 
\usepackage{algorithm,algorithmic}
\usepackage[margin=2cm]{geometry}
\usepackage{comment}

\theoremstyle{definition}
\theoremstyle{remark}

\usepackage{soul}

\usepackage{mathrsfs}
\usepackage{authblk}

\title{Impact of Model Size on Fine-tuned LLM Performance in Data-to-Text Generation: A State-of-the-Art Investigation}
\author[1]{Joy Mahapatra \thanks{joymahapatra90@gmail.com}}
\author[1]{Utpal Garain \thanks{utpal@isical.ac.in}}
\affil[1]{Indian Statistical Institute Kolkata}

\date{}

\begin{document}
\maketitle

\begin{abstract}
    Data-to-text (D2T) generation aims to generate human-readable text from semi-structured data, such as tables and graphs. 
    The recent success of D2T is largely attributed to advancements in LLMs.
    Despite the success of LLMs, no research has been conducted to illustrate the impact of model size on the performance of fine-tuned LLMs for D2T tasks.
    D2T model performance is typically assessed based on three key qualities: \textit{readability} (indicates fluency and coherence), \textit{informativeness} (measures content similarity), and \textit{faithfulness} (assesses consistency of factual information).
    It is currently uncertain whether increasing the size of LLMs effectively improves performance in D2T tasks across these three qualities.
    The objective of this study is to investigate the performance of fine-tuned LLMs in D2T tasks in terms of model size.
    Through extensive comparative analysis, we aim to elucidate both the advantages and limitations of scaling model sizes across five widely used D2T datasets (E2E, ViGGo, WikiTableText, DART, and WebNLG) and twelve state-of-the-art LLMs with varying sizes from five different LLM families (T5, BART, OPT, BLOOM, and Llama 2). 
    To comprehensively cover all the three essential qualities of D2T models, we incorporate six widely recognized automatic metrics---\textsc{BLEU}, \textsc{METEOR}, \textsc{BERTScore}, \textsc{MoverScore}, \textsc{Parent}, and \textsc{BARTScore}.
    We also provide an in-depth analysis of LLM performance concerning model size in the presence of source-reference divergence, a critical aspect of D2T tasks.
    Our investigation reveals that increasing LLM size enhances \textit{readability} and \textit{informativeness} in D2T tasks, but larger (in terms of size) LLMs may sacrifice \textit{faithfulness}.
    Moreover, small-sized LLMs show more resilience than larger ones when source-reference divergence is present.
\end{abstract}

\section{Introduction}
\label{sec:introduction}
Since their inception, large language models (LLMs) have emerged as dominant models in text generations, surpassing their predecessor models in terms of generality and performance~\citep{raffel2020exploring,brown2020language,chowdhery2023palm,touvron2023llama}.
Current LLMs have demonstrated exceptional performance in a multitude of text generation tasks, including but not limited to machine translation, automatic summarization, and text simplification~\citep{lewis2020bart,zhang2022opt,taori2023stanford}.
The remarkable effectiveness of LLMs in these tasks has led to a clear trend of increasing their sizes (numbers of model parameters) in existing LLMs to further enhance performance~\citep{petroni2019language,roberts2020how}.
Consequently, there has been an emergence of numerous new LLM families~\citep{touvron2023llama,chowdhery2023palm,scao2022bloom} characterized by substantially larger sizes, which emphasizes the ongoing progress within the field of text generation.

\begin{figure}[ht]
    \centering
    \resizebox{\textwidth}{!}{
    \includegraphics{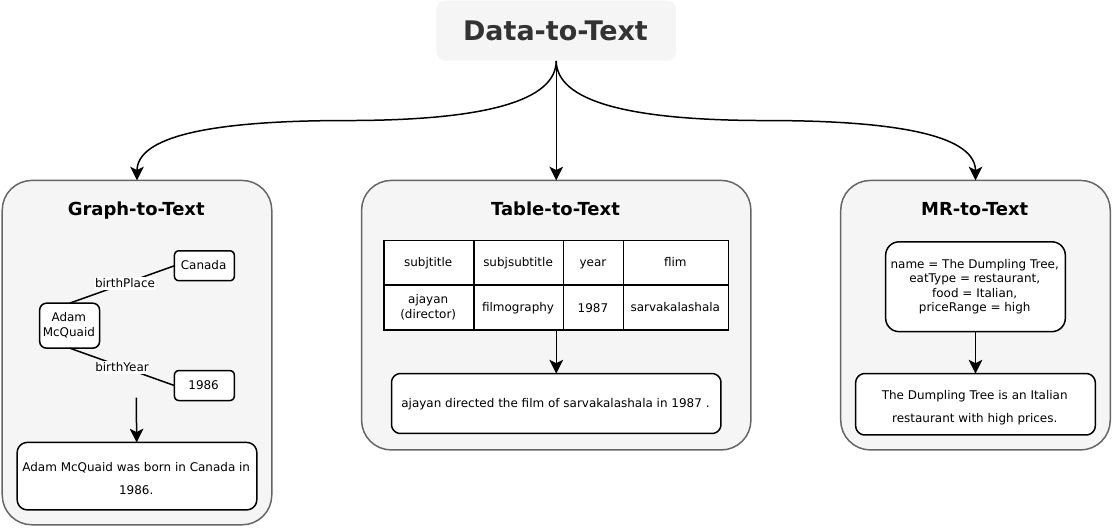}}
    \caption{Overview of data-to-text (D2T) generation with three major types: graph-to-text (left), table-to-text (middle), MR (meaning representation)-to-text (right).}
    \label{fig:d2t}
\end{figure}

Data-to-text (D2T) generation~\citep{lin2024survey}, an essential facet of text generation, where goal is to produce human-readable text from semi-structured data sources, including slot-value paired meaning representations (MR)~\citep{novikova2017e2e}, tables~\citep{bao2018table}, or graphs~\citep{nan2021dart}.
Based on the source data types, three major types of D2T tasks are shown in Figure~\ref{fig:d2t}.
D2T has diverse applications across various domains, spanning dialogue generation~\citep{wen2015semantically}, sports reporting~\citep{wiseman2017challenges}, weather forecasting~\citep{belz2008automatic}, business intelligence~\citep{gatt2018survey}, and many more.
In critical fields like healthcare, D2T plays a vital role in automatic diagnostic reporting~\citep{hommes2019personalized,yermakov2021biomedical}, underscoring its relevance in critical decision-making.
The recent success of D2T tasks can be largely attributed to the advancement of LLMs, which are leveraged through several effective inference and training methods such as parameter-efficient fine-tuning~\citep{li2021prefix,lester2021power,dettmers2023qlora}.
The synergy between LLMs and D2T generation marks a significant advancement~\citep{li2024unifying,jing2024stylized,kasner2024reference} across various domains and critical safety applications.

\begin{figure}[ht]
    \centering
    \includegraphics[scale=0.7]{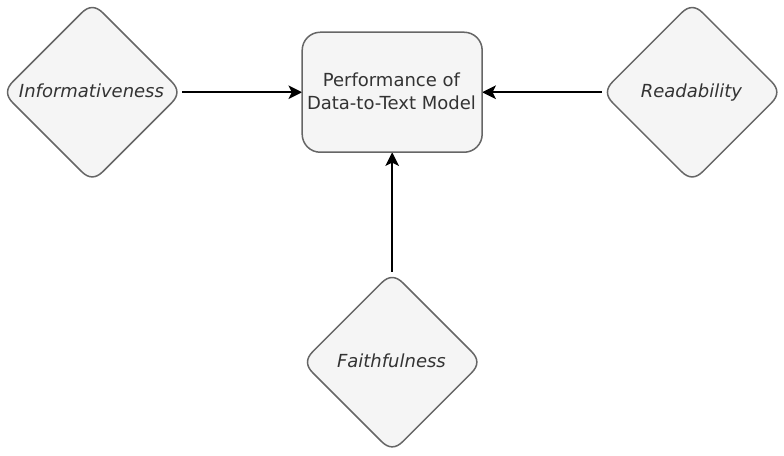}
    \caption{Three key qualities to assess the performance of a D2T model are: \textit{readability} (focusing on fluency and coherence), \textit{informativeness} (evaluating the ability to generate essential content), and \textit{faithfulness} (indicating the consistency of the generated text by measuring the presence of irrelevant facts).}
    \label{fig:three_qualities}
\end{figure}

Despite advancements in LLMs for D2T tasks, there exists a siggnificant gap in analyzing the impact of model size on the performance of fine-tuned LLMs as D2T models.
The assessment of D2T models commonly revolves around three fundamental performance qualities: \textit{readability}, \textit{informativeness}, and \textit{faithfulness} (see Figure~\ref{fig:three_qualities}).
\textit{Readability}~\citep{chen2020logic2text,li2022faithfulness} mainly concerns the fluency and coherence of generated text from the model, whereas \textit{informativeness}~\citep{shen2019pragmatically,li2022faithfulness} determines whether the D2T model is effective in terms of transforming essential information from given data to generated text. 
\textit{Faithfulness}~\citep{tian2019sticking,wang2020towards} is an important performance indicator for D2T, assessing whether the generated text presents any incorrect or irrelevant facts w.r.t. the given source data.
Considering the huge potential of LLMs in various critical D2T domains~\citep{hommes2019personalized,pauws2019making}, it is crucial to understand the impact of model size over fine-tuned LLMs in terms of D2T performance.
However, almost no existing literature has investigated the impact of model size on the performance of fine-tuned LLMs across all three performance qualities of D2T.

This paper aims to address the research gap by providing a comparative analysis to demonstrate the impact of model size (number of parameters) on the performance of fine-tuned LLMs for D2T tasks. 
We conduct the comprehensive performance comparison of twelve LLMs with varying sizes, drawn from five widely used LLM families: BART~\citep{lewis2020bart}, T5~\citep{raffel2020exploring}, OPT~\citep{zhang2022opt}, BLOOM~\citep{scao2022bloom}, and Llama 2~\citep{touvron2023llama}.
We measure the performances of LLMs on D2T tasks across three key qualities of D2T model---\textit{readability}, \textit{informativeness}, and \textit{faithfulness}.
To cover a broad spectrum, we include three primary types of D2T tasks---table-to-text, graph-to-text, and MR-to-text.
All of these D2T are depicted in Figure~\ref{fig:d2t}.
We incorporate five state-of-the-art datasets in our experiments: E2E~\citep{novikova2017e2e} (MR-to-text), ViGGo~\citep{juraska2019viggo} (MR-to-text), WikiTableText~\citep{bao2018table} (table-to-text), DART~\citep{nan2021dart} (graph-to-text), and WebNLG~\citep{gardent2017webnlg} (graph-to-text).
Through comparative analyses, we aim to investigate both the advantages and limitations of scaling up the parameters in LLMs (or size of LLMs) within the domain of D2T.
Additionally, we provide insights into the performance of LLMs in terms of their size with considering the factual/information divergence between the source and reference data---known as source-reference divergence.
Through considering source-reference divergence, we aim to assess LLMs' generalization capability in D2T tasks and evaluate if increasing LLM size contributes to quality generation when references diverge from the source data in D2T.

The organization of the remainder of the paper is as follows.
Section~\ref{sec:aim_motivation} outlines the research questions and motivations behind this paper.
Prevalent related work concerning D2T and application of LLMs is presented in Section~\ref{sec:related_work}.
Section~\ref{sec:preliminaries} offers a concise overview of conditional text generation, (large) language models, D2T, and source-reference divergence. 
All details about the models, datasets, and experimental settings are provided in Section~\ref{sec:models_datasets_settings}.
In Section~\ref{sec:evaluation_results}, we present a comparative performance evaluation of LLMs with different model sizes for D2T tasks. Section~\ref{sec:effect_divergence} analyzes the impact of model size on LLM performance in the presence of source-reference divergence in D2T tasks. 
Section~\ref{sec:case_studies} includes case studies to illustrate the outcomes of LLMs more clearly.
Finally, we conclude our findings in Section~\ref{sec:conclusion}.

\section{Research Questions and Motivations}
\label{sec:aim_motivation}
This paper aims to empirically address the following research questions regarding the performance of fine-tuned LLMs for D2T, focusing impact of model sizes (i.e., parameter counts) across the three key qualities of \textit{readability}, \textit{informativeness}, and \textit{faithfulness}.
\begin{enumerate}
    \item What are the impacts of model size within a family of fine-tuned LLMs on the performance of data-to-text (D2T) tasks, in terms of the \textit{readability}, \textit{informativeness}, and \textit{faithfulness}?
    \item Do larger LLM families (such as OPT, BLOOM, Llama 2, etc.) convincingly outperform smaller LLM families (such as BART, T5, etc.) in terms of D2T task performance?
    \item Does the presence of source-reference divergence influence the performance of LLMs for D2T tasks? If so, does increasing the model size of LLM aid in mitigating the effects of source-reference divergence?
\end{enumerate}

The motivation behind our objectives stems from several key considerations.
Firstly, the recent trend of increasing number of parameters in LLMs comes with significant computational costs and longer inference times~\citep{zheng2023response,li2024sustainable,jiang2024preventing}.
If this escalation in model size does not yield substantial improvements~\citep{aghajanyan2021intrinsic,tulchinskii2023intrinsic} in D2T, it raises questions about the necessity of bearing such computational burdens~\citep{luccioni2023estimating,faiz2023llmcarbon}.
Secondly, despite the widespread use of LLMs, their performance in D2T tasks has not been comprehensively explored.
This research gap hinders the effective application of LLMs in D2T scenarios.
Lastly, LLMs are known for their generalization capabilities~\citep{ge2023openagi,zhao2024expel,yang2024unveiling}, as evidenced by their success in various generative text tasks.
Therefore, it is pertinent to investigate whether this generalization extends to enhancing LLM performance for D2T tasks in the presence of source-reference divergence.

\section{Related Work}
\label{sec:related_work}
Traditionally, D2T models have relied on rule-based methodologies~\citep{reiter1997building}, meticulously crafted by domain experts. However, these rule-based approaches have been hindered by two primary challenges: scalability and the scarcity of domain experts~\citep{gatt2018survey,angeli2010simple}. As the demand for D2T models continues to rise across various applications~\citep{lu2011probabilistic,gatt2018survey}, there has been a notable shift towards the development of automatic D2T models. 
Statistical n-gram based language models~\citep{brown1992class,bickel2005predicting,pauls2011faster} laid the foundation for probabilistic D2T models~\citep{chen2008learning,angeli2010simple,lu2011probabilistic}. 
While these initial probabilistic models have gained traction due to their scalability, they still struggle to emulate human-like text generation for D2T tasks~\citep{belz2006comparing,reiter2018structured}.
The emergence of recurrent neural network-based (RNN) language models~\citep{mikolov2010recurrent} marked a significant milestone in the realm of D2T tasks~\citep{lebret2016neural}, enabling D2T models to exhibit extreme fluency.
With the introduction of encoder-decoder framework-based sequence-to-sequence neural networks~\citep{sutskever2014sequence}, D2T generation experienced a surge in popularity compared to existing methods~\citep{graves2013speech}. 
These models~\citep{wen2015semantically,shang2015neural} excel in \textit{readability}, representing a significant advancement not only in D2T but also in several text generation tasks.
Nevertheless, these RNN-based encoder-decoder D2T models still struggle to ensure the \textit{faithfulness} and \textit{informativeness} of generated text~\citep{wiseman2017challenges}.
With the integration of attention mechanisms~\citep{bahdanau2015neural,luong2015effective,xu2015show} into deep encoder-decoder networks and the introduction of pre-training/fine-tuning mechanisms~\citep{howard2018universal}, numerous D2T models~\citep{mei2016what,liu2018table,sha2018order,nema2018generating,budzianowski2018multiwoz,juraska2018deep,nie2019encoder,puduppully2019data,gong2019table} have emerged.
The advent of the transformer models~\citep{vaswani2017attention} has led to their widespread adoption across various NLP tasks, including D2T generation.
Almost all recent language models, built upon the transformer architecture, play pivotal roles in text generation tasks, including D2T~\citep{erdem2022neural,ji2023survey,lin2024survey}.
These transformer-based pre-trained language models serve as foundational models for D2T tasks, benefiting from pre-training on large text datasets using self-supervised learning~\citep{taylor1953cloze,bickel2005predicting}.
Through fine-tuning strategies, these language models have gained immense popularity in D2T tasks~\citep{devlin2019bert,lewis2020bart,radford2019language,brown2020language}.
D2T models built upon these language models often achieve higher rankings compared to earlier D2T models~\citep{ge2023openagi,zhang2024can}.
Utilizing language models in D2T tasks also streamlines the training process, as they require only a small amount of task-specific data for fine-tuning~\citep{erdem2022neural}. 
While language model-led D2T systems show improvements in \textit{informativeness}, they still struggle to generate \textit{faithful} content.

Several recent studies~\citep{petroni2019language,heinzerling2021language,roberts2020how} on language models, suggest that the parameters within these models are responsible for encoding knowledge.
To enhance the knowledge capacity of these models, recent efforts have significantly increased the number of parameters~\citep{raffel2020exploring,scao2022bloom,touvron2023llama}.
These large, parameter-rich language models (i.e., LLMs) demonstrate remarkable performance across various generative tasks, such as text generation, and discriminative tasks, like text classification~\citep{devlin2019bert,lewis2020bart,radford2019language}.
Even in few-shot and zero-shot training setups~\citep{brown2020language}, these LLMs consistently excel in numerous NLP tasks.
The trend of augmenting parameters within LLMs has significantly shaped the landscape of text generation, with LLMs significantly outperforming earlier deep learning-based models.
This phenomenon has also impacted D2T tasks, as the majority of recent D2T models utilize LLMs~\citep{kasner2024reference,li2024unifying,jing2024stylized,lorandi2024high}.
Recent advancements in parameter-efficient fine-tuning approaches for LLMs, such as prefix-tuning~\citep{li2021prefix}, P-tuning~\citep{liu2022p}, and prompt tuning~\citep{lester2021power}, along with popular adapter-based fine-tuning methods like LoRA~\citep{hu2022lora}, have made LLMs more suitable for D2T tasks.
Given that LLMs are pre-trained on vast text corpora~\citep{raffel2020exploring,touvron2023llama}, they excel in capturing rich contextual information, making them well-suited for generating text in D2T scenarios.

Despite the widespread application of LLMs in D2T, several important questions persist regarding their performance, particularly considering their sizes (numbers of parameters)~\citep{luccioni2023estimating,jiang2024preventing}.
Achieving human-like \textit{readability} in generated text remains a challenge for LLMs in D2T~\citep{jing2024stylized}.
Regarding \textit{informativeness}, a crucial quality of D2T models, LLMs may overlook essential information when generating from non-textual data due to underlying biases stemming from their large number of parameters~\citep{wu2023style}.
Recent studies highlight that even with an increase in parameters, the effective or extrinsic parameters of LLMs remain low~\citep{aghajanyan2021intrinsic,tulchinskii2023intrinsic}.
Concerns also persist regarding the \textit{faithfulness} of LLMs in D2T~\citep{ji2023survey}.
Recent investigations have sought to highlight the limitations of LLMs as D2T models.
However, these studies have primarily consisted of surveys discussing the progress of D2T tasks~\citep{lin2024survey} or outlining D2T models focusing on single key attributes~\citep{gatt2018survey,erdem2022neural,ji2023survey}. 
None of these studies offers a comprehensive analysis of D2T models considering all three essential qualities (\textit{readability}, \textit{informativeness}, and \textit{faithfulness}), particularly in relation to LLMs.
Our study aims to address this research gap.

\section{Preliminaries}
\label{sec:preliminaries}
\subsection{Conditional Text Generation}
\label{subsec:condition_generation}
Text generation or natural language generation~\citep{lu2018neural,erdem2022neural}, both fundamental and crucial task in NLP, encompasses various applications such as machine translation~\citep{gatt2018survey}, D2T~\citep{ye2020variational}, dialogue generation~\citep{wen2015semantically}, image captioning~\citep{karpathy2017deep}, summarization~\citep{see2017get,islam2023tackling}, story generation~\citep{fan2018hierarchical}, report writing~\citep{lu2011probabilistic,lin2024survey}, etc. 
These text generation tasks can be broadly categorized into two main types: open-ended text generation and conditional (or controlled) text generation~\citep{holtzman2018learning,li2022faithfulness}.
In open-ended text generation, models produce text without any constraint from prior source information.
For illustration, generating story narratives after a given context is an open-ended generation~\citep{fan2018hierarchical}.
Completing text from a text fragment is also an example of open-ended generation.
On the other hand, in conditional text generation, the model generates text based on a constraint given through prior source information and instructions.
Examples of conditional text generation tasks include automatic summarization, machine translation, and task-oriented dialogue generation, all of which rely on specific input (source) conditions for text generation.
D2T falls within conditional text generation~\citep{puduppully2019data,upadhyay2023cbr}, as it involves generating text from structured data inputs, further emphasizing its significance in this category.
Conditional text generation tasks are always represented through conditional probability distributions~\citep{graves2013speech,gatt2018survey,erdem2022neural}.
To generate a text, $w=w_1,w_2,w_3, \dots, w_n$ (where $w_i$ is $i$-th word/token, remember $w_0$ means empty word) from a source, $s$, we can be explicitly written as follow:
\begin{center}
    \begin{align*}
        w_i \sim  \left.Pr\left(\cdot \Big\vert {\color{black} \underbrace{s}_{\text{Given source}}}, {\color{black} \underbrace{w_1,w_2,w_3, \dots, w_{i-1}}_{\text{Previous generated context}}}\right)\right\vert_{i=1}^{n-1}
    \end{align*}
\end{center}

\begin{figure}[ht]
    \centering
    \includegraphics[scale=0.8]{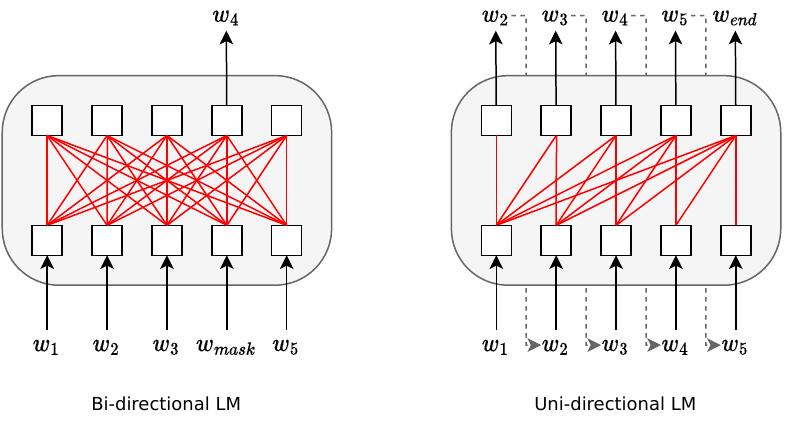}
    \caption{Two prevalent types of language models based on transformer~\citep{vaswani2017attention} architecture are depicted here: bidirectional and unidirectional language models. The red lines represent attention mechanisms.}
    \label{fig:lm_types}
\end{figure}

\subsection{(Large) Language Model}
\label{subsec:llm}
Most of the recent advancement of generative text generation tasks are direct outcomes of predictive language modelling~\citep{pauls2011faster,mikolov2010recurrent,devlin2019bert,radford2019language}.
With the advent of transformers~\citep{vaswani2017attention} and pre-training paradigm~\citep{howard2018universal}, language models have demonstrated unparalleled dominance across various NLG tasks.
Primarily there are mainly two type of language modelling (Figure~\ref{fig:lm_types})--- bidirectional and uni-directional language modelling~\citep{lu2018neural}.
In bidirectional language model, both left and right contexts are considered in predicting a context.
Masked language modelling~\citep{taylor1953cloze}, where the objective is to predict mask word in a text, paved the path for bidirectional language model.
With the appearance of BERT~\citep{devlin2019bert}, bi-directional language model show performance enhancement both in terms of semantic and contextual informativeness in several NLP task.
Objective of language model is to predict masked words ($w_i$) from a given masked context ($w_{1} \dots w_{i-1} w_{\text{mask}} w_{i+1} \dots w_{n}$), as follows:
\begin{align*}
    w_i \leftarrow \underset{w}{\text{argmax}}\ p(w \vert w_{1} \dots w_{i-1} w_{\text{mask}} w_{i+1} \dots w_{n})
\end{align*}

On the other hand, unidirectional language models predicts target words from one direction, mostly left-to-right directional are followed~\citep{radford2019language,brown2020language,raffel2020exploring}.
Earlier, unidirectional language model seems to have less powerful compare to the bi-directional language model.
However several recent studies~\citep{li2021prefix,liu2022p} have shown that unidirectional language models can also be powerful as bi-direction language model in terms of language understanding.
Currently most of the popular language models are uni-directional language models.
Objective of unidirectional language model is to predict next words ($w_n$) from a given prior contexts ($w_1w_2 \dots w_{n-1}$),
\begin{align*}
    w_n \leftarrow \underset{w}{\text{argmax}}\ p(w \vert w_1w_2 \dots w_{n-1})
\end{align*}

\begin{figure}[ht]
    \centering
    \resizebox{\textwidth}{!}{
    \includegraphics{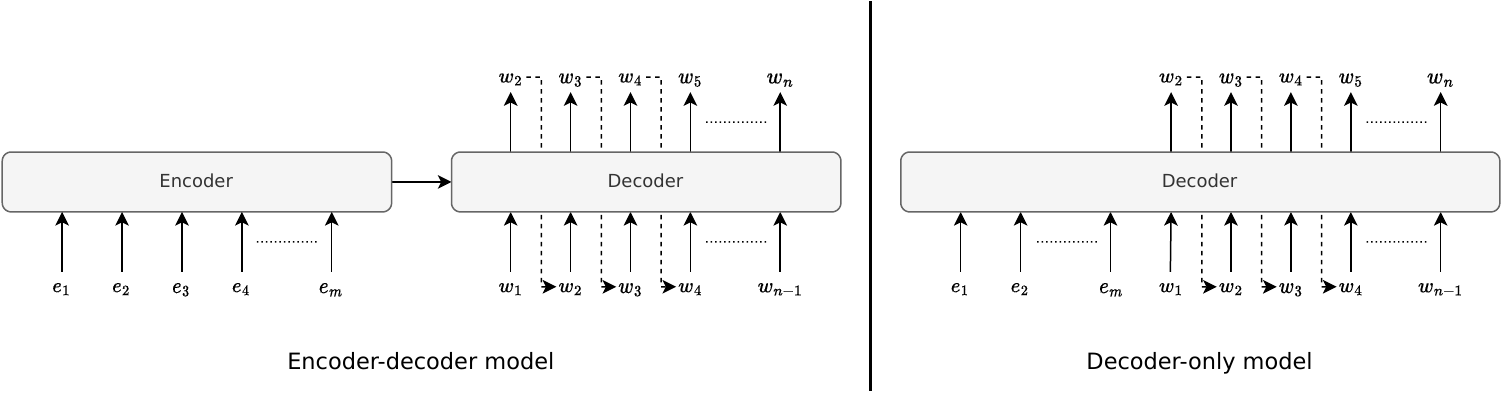}}
    \caption{Two of the most popular architectures used for implementing language models: the encoder-decoder architecture (left) and the decoder-only architecture (right). The sequence $e_1e_2\dots e_m$ represents the source input of size $m$ (number of words), while the corresponding textual output is denoted by $w_1w_2\dots w_n$ of size $n$.}
    \label{fig:seq2seq}
\end{figure}

Two primary architectures are commonly used for implementing language models: encoder-decoder architecture~\citep{sutskever2014sequence,vinyals2016order} and decoder-only architecture~\citep{radford2019language,brown2020language} (Figure~\ref{fig:seq2seq}). 
In the encoder-decoder architecture, the encoder part encodes an input sequence, and the decoder generates a new sequence from the encoded representation. On the other hand, in the decoder-only architecture, the decoder auto-regressively completes a sequence from a given context.

In contemporary times, large language models (LLMs) have become omnipresent in nearly all NLP tasks, owing to their remarkable performance and versatility~\citep{devlin2019bert,brown2020language,touvron2023llama}.
LLMs stand out from previous language models primarily due to two factors: their enormous size~\citep{chowdhery2023palm}, characterized by a vast number of underlying parameters, and their extensive self-supervised pre-training on massive text corpora~\citep{raffel2020exploring,penedo2023refinedweb}.
Compared to earlier language models~\citep{graves2013speech,mikolov2010recurrent,peters2018deep}, contemporary LLMs boast billions, and even trillions, of parameters, enabling them to achieve exceptional language understanding~\citep{roberts2020how,aghajanyan2021intrinsic}.
Additionally, LLMs demonstrate impressive results through supervised fine-tuning, also known as instruction tuning~\citep{si2023empirical,zhang2023instruction}, and training aided by reinforcement learning techniques~\citep{fernandes2023bridging,ouyang2022training}, incorporating human-like feedback through preference models~\citep{touvron2023llama,chowdhery2023palm}.

\subsection{Data-to-Text (D2T) Generation}
\label{subsec:d2t}
The aim of Data-to-Text generation (D2T)~\citep{lin2024survey} is to convert non-textual, semi-structured source data, such as tables, graphs, or slot-value pairs (meaning representation, MR), into human-readable text output.
D2T encompasses various types based on source representation, including graph-to-text, table-to-text, and MR-to-text (Figure~\ref{fig:d2t}).
In graph-to-text~\citep{gardent2017webnlg,nan2021dart}, structured graph data or knowledge triple sets are transformed into coherent narratives, while table-to-text~\citep{liu2018table,bao2018table,gong2019table} involves converting tabular data into fluent textual descriptions.
MR-to-text focuses on generating textual reports from slot-value pairs, also known as Meaning Representation (MR)~\citep{novikova2017e2e,juraska2019viggo,sha2018order,kedzie2020controllable}.
While D2T shares similarities with other text generation tasks like machine translation and summarization, its primary distinction lies in the use of semi-structured non-textual data as input~\citep{ji2023survey}.
As the demand for automated text generation from semi-structured data grows, D2T remains a crucial research area with promising implications for enhancing data accessibility and decision-making processes~\citep{gatt2018survey}.
Numerous predictive models have been developed for D2T generation, with recent advancements largely attributed to the use of predictive language models~\citep{pauls2011faster,mikolov2010recurrent,devlin2019bert,radford2019language}.
These models, utilizing recurrent neural networks (RNNs)~\citealp{hochreiter1997long,graves2013speech,mikolov2010recurrent} and transformer architecture~\citep{vaswani2017attention}, exhibit significant potential across various D2T tasks.
Transformers have attracted significant attention and demonstrated remarkable success in D2T applications~\citep{su2021plan,wang2020towards}.

D2T is typically learned from a dataset consisting of a set of source-reference pairs, $\mathcal{D} = \left.{\{s, r\}}\right \vert_{\in \mathcal{D}}$, where $s$ is semi-structured source data with its corresponding reference text $r = r_{1}r_{2},\dots,r_{|r|}$, and each word belong to vocabulary set $\mathcal{V}$, i.e., $r_{i} \in \mathcal{V}$~\citep{nema2018generating,nie2019encoder,nie2019simple}.
Any predictive D2T model involves three crucial phases: Modelling, Training, and Inference.
The Modelling phase involves selecting the structure of the predictive D2T model, denoted as $\mathcal{M}_{\theta}$ and parameterized by $\theta$, i.e., $\mathcal{V}^{*} \sim \mathcal{M}_{\theta}(x)$. 
These predictive D2T models~\citep{sha2018order,liu2018table,dusek2018findings,gong2019table} often follow either encoder-decoder or decoder-only paradigms, with transformer-based architectures being commonly used for both encoder and decoder components.
In training or learning phase, the predictive D2T model $\mathcal{M}_{\theta}$ is trained with the dataset $\mathcal{D}$, typically using maximum likelihood estimation~\citep{gkatzia2016content,holtzman2018learning,puduppully2021data,luo2023data} strategies (equation~\ref{eq:training}).

\begin{align}
    \theta^{*} \leftarrow \underset{\theta}{\text{argmax}}\sum\limits_{(s,r) \in \mathcal{D}}\mathcal{L}(\mathcal{M}_\theta(s), r) \label{eq:training}
\end{align}

Here, $\mathcal{L}$ represents the underlying loss function, which is commonly taken as the cross-entropy loss.
During inference, a trained predictive D2T model, $\mathcal{M}_{\theta^*}$, undergoes the text generation process to produce output $g$ (an approximation of $r$) from a given source data, $s$ (equation~\ref{eq:decoding}).
At this stage, various decoding strategies~\citep{fan2018hierarchical,holtzman2020curious} are applied to the D2T model.
Some popular decoding strategies are beam search decoding, greedy search, nucleus sampling~\citep{holtzman2020curious}, Top-$K$~\citep{fan2018hierarchical}, etc.

\begin{align}
    g \leftarrow \underset{\mathcal{V}^{*}}{\text{decoding}}\ \mathcal{M}_{\theta^*}(s) \label{eq:decoding}
\end{align}

\subsection{Source-Reference Divergence}
\label{subsec:divergence}
\begin{figure}[ht]
    \centering
    \resizebox{0.7\textwidth}{!}{
    \includegraphics{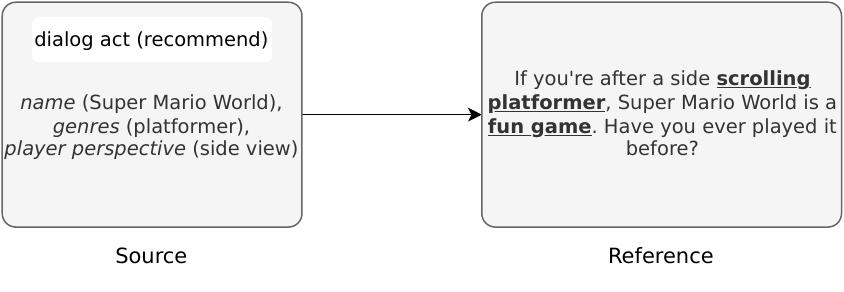}}
    \caption{An example of source-reference divergence, taken from the WikiTableText~\citep{bao2018table} dataset. The reference text includes two additional facts (bolded and underlined) - `scrolling performer' and `fun game' - that are absent in the corresponding source data.}
    \label{fig:divergence}
\end{figure}

Previous studies~\citep{dhingra2019handling,tian2019sticking,islam2023tackling} have noted that across various D2T tasks~\citep{lebret2016neural}, there often exists a discrepancy between the information or facts present in the source data ($s$) and the corresponding reference data ($r$).
This disparity is known as source-reference divergence~\citep{dhingra2019handling,li2022faithfulness}.
An illustrative example of such divergence is depicted in Figure~\ref{fig:divergence}.

The primary causes of such divergence stem from the processes of data collection and the inherent nature of the tasks~\citep{lebret2016neural,gardent2017webnlg,dusek2018findings}.
In many cases, source and reference texts are drawn from disparate origins, resulting in a weak factual correlation between them.
Additionally, D2T datasets are often curated by multiple human annotators who interpret the source data before generating corresponding textual references~\citep{nie2019simple}.
Each annotator brings their unique background knowledge and comprehension abilities to the task, leading to generated references that may include additional or misaligned facts. 
Another significant factor contributing to source-reference divergence in D2T tasks is the inherent nature of D2T tasks~\citep{li2022faithfulness}.
Source data of D2T task are typically more compact and specific, whereas references tend to be more generalized.
Consequently, source-reference divergence is common and challenging to eliminate entirely, particularly in D2T contexts~\citep{dhingra2019handling,nie2019simple}. 
Therefore, evaluating D2T model performance in the presence of such divergence is of utmost importance.

\section{Models, Datasets and Experimental Settings}
\label{sec:models_datasets_settings}
In this section, we describe the large language models (LLMs) and data-to-text (D2T) datasets employed in our subsequent experiments.
Additionally, we include a subsection outlining the experimental settings, where we provide detailed information regarding the experimental setup.

\subsection{Models}
\label{subsec:models}
\begin{figure}[ht]
    \centering
    \includegraphics[scale=0.7]{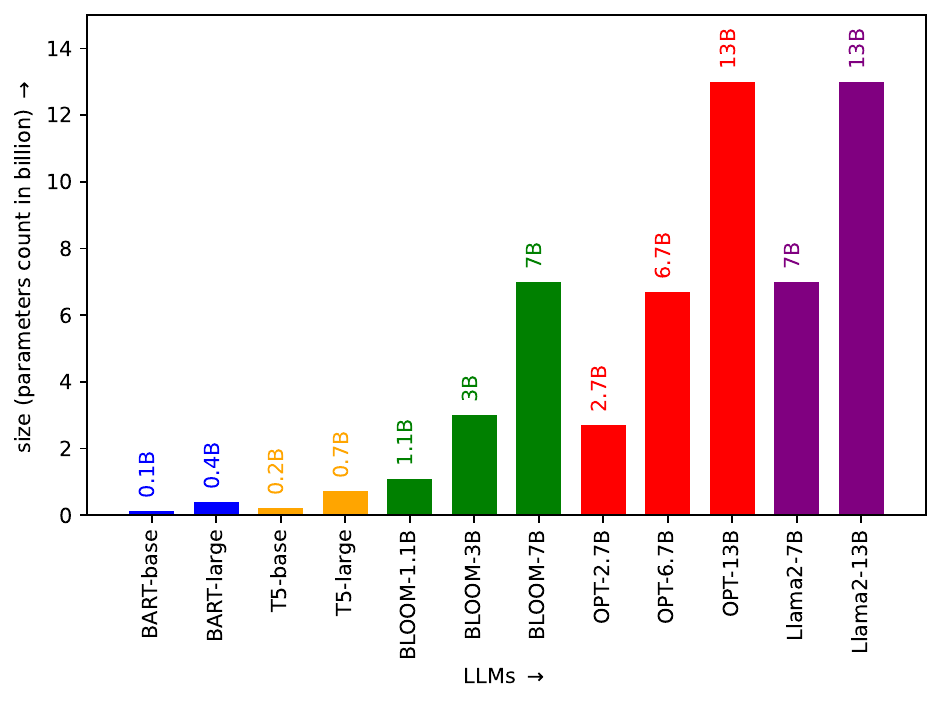}
    \caption{Sizes (number of parameters) of the 12 LLMs from five LLM families included in our experiments. BART, T5, BLOOM, OPT, and Llama 2 families are indicated with blue, orange, green, red, and purple respectively.}
    \label{fig:lm_expansion}
\end{figure}

We include 12 popular LLMs from five widely recognized LLM families (BART, T5, BLOOM, OPT, and Llama 2), all of which are widely utilized in the text generation field.
We select LLMs of varying model sizes from each of the five model families. 
Considering the model sizes, among these five LLM families, we categorize BART and T5 as families of smaller LLMs, while the remaining three are families of larger LLMs.
Figure~\ref{fig:lm_expansion} depicts all 12 incorporated LLMs along with their sizes, i.e., the number of parameters. All these LLMs are sourced from Hugging Face's Model Hub~\citep{wolf2020transformers}.

\paragraph{BART.}
BART~\citep{lewis2020bart} is an encoder-decoder architecture-based language model family.
The encoder encodes the given context through bidirectional language modeling, while the decoder performs auto-regressive generation.
BART is pre-trained using a self-supervised approach involving text denoising, which includes methods like token masking, token deletion, text infilling, etc.
In our experiments, we use two models of the BART family, which differ in terms of their sizes: \textsf{BART-base} (139 million parameters) and \textsf{BART-large} (406 million parameters).

\paragraph{T5.}
Similar to BART, T5~\citep{raffel2020exploring} follows the encoder-decoder paradigm. 
However, the main motivation behind T5 is to unify several NLP tasks and train them together as a single task. 
T5 is pre-trained using BERT-like span-corruption techniques~\citep{devlin2019bert} on a large crawled dataset called C4 (Colossal Clean Crawled Corpus)~\citep{raffel2020exploring}. 
T5 has demonstrated state-of-the-art results on several natural language understanding tasks. 
In our experiments, we utilize two models of T5 family with different model sizes: \textsf{T5-base} (223 million parameters) and \textsf{T5-large} (738 million parameters).

\paragraph{BLOOM.}
BLOOM~\citep{scao2022bloom} is a publicly available decoder-only language model family trained on a large dataset---ROOT dataset, which includes 46 natural languages and 13 programming languages. 
Similar to BART and T5, BLOOM is based on a transformer architecture. 
In our study, we consider three different sizes for BLOOM LLMs: \textsf{BLOOM-1B} (1 billion parameters), \textsf{BLOOM-2.7B} (2.7 billion parameters), and \textsf{BLOOM-6.7B} (6.7 billion parameters).

\paragraph{OPT.}
Like BLOOM, OPT~\citep{zhang2022opt} is an auto-regressive decoder-based language model family that utilizes a transformer decoder architecture. 
OPT is pre-trained on three datasets: the RoBERTa dataset~\citep{liu2019roberta}, the Pile, and PushShift.io Reddit~\citep{zhang2022opt}. 
These corpora were previously collected or filtered.
OPT has demonstrated performance comparable to popular GPT models on several state-of-the-art benchmarks. 
Unlike GPT~\citep{brown2020language}, OPT is fully publicly available.
In this paper, we use three OPT models with different model sizes: \textsf{OPT-2.7B} (2.7 billion parameters), \textsf{OPT-6.7B} (6.7 billion parameters), and \textsf{OPT-13B} (13 billion parameters).

\paragraph{Llama 2.}
Llama 2 family~\citep{touvron2023llama} is the successor of Llama from~\href{https://ai.meta.com/}{Meta AI}.
It is a decoder-based LLM trained through three stages: traditional pre-training, supervised fine-tuning, and human feedback endorsed by reinforcement learning. 
Llama 2 is pre-trained with almost 2 trillion tokens and is ranked higher in terms of human safety and usability compared to other LLMs. 
It shows comparable results with several proprietary LLMs like ChatGPT. 
In our experiments, we utilize two Llama 2 models with different sizes: \textsf{Llama2-6B} (6 billion parameters) and \textsf{Llama2-13B} (13 billion parameters).

\subsection{Datasets}
We included five state-of-the-art D2T datasets covering all three major types of D2T tasks: E2E and ViGGO for MR-to-text, WikiTableText for table-to-text, and DART and WebNLG for graph-to-text.
Table~\ref{tab:dataset_statistics} presents key statistics for all five D2T datasets.
All these dataset are downloaded from \citet{kasner2023tabgenie} and \citet{wolf2020transformers}.

\paragraph{E2E dataset.}
This dataset contains information from the restaurant domain and is frequently utilized in D2T tasks~\citep{novikova2017e2e,dusek2018findings}.
It consists of input data in the form of slot-value pairs, where each slot represents an attribute and its corresponding value indicates the attribute's realization.
Known for its lexical richness and diverse syntactical structures, this dataset is highly esteemed in D2T research.
Comprising over 2 million tokens (including both references and source) and almost 37K data-text pairs, with approximately 4.5K unique words, it provides abundant data for both training and evaluation purposes.

\paragraph{ViGGo dataset.}
The ViGGO dataset~\citep{juraska2018deep}, focusing on the video game domain, is ideal for training open-domain D2T models. 
With a diverse range of dialogue acts and forms drawn from nearly a hundred video games, ViGGO offers extensive coverage and high text diversity. 
Containing around 7K instances, ViGGO exhibits greater lexical richness compared to other datasets like E2E. 
Each meaning representation in ViGGO corresponds to one of nine distinct conversational dialogue acts, providing nuanced dialogue interactions. 
Overall, ViGGO excels in prioritizing lexical richness and text diversity, making it a valuable resource for D2T tasks.

\paragraph{WikiTableText dataset.} 
WikiTableText~\citep{bao2018table} is designed for the table-to-text form of Data-to-Text (D2T) task. 
Unlike the E2E and ViGGO datasets, it belongs to the open-domain category. It consists of numerous domain Wikipedia tables, randomly selected and manually annotated to generate corresponding text.
With a total of $\sim 13$K data-text pairs, each pair contains an average of 13.9 tokens.

\begin{table}[ht]
    \centering
    \resizebox{\textwidth}{!}{
    \begin{tabular}{@{}cccccccccc@{}}
    \toprule
    \multirow{2}{*}{dataset} & \multirow{2}{*}{D2T types} & \multirow{2}{*}{domain} & dataset size & \multicolumn{3}{c}{source (linearized)} & \multicolumn{3}{c}{reference} \\ \cmidrule(lr){5-7} \cmidrule(lr){8-10} 
     &  &  & (\# instances) & average length & unique tokens & total tokens & average length & unique tokens & total tokens \\ \midrule
    E2E & mr-to-text & closed & 36,856 & 27.3 & 125 & 1M & 20.8 & 4.5K & 885K\\
    ViGGo & mr-to-text & closed & 6,900 & 29.9 & 618 & 206K & 21.5 & 4.4K & 148K \\
    WikiTableText & table-to-text & open & 13,318 & 35.2 & 29K & 469K & 13.9 & 24K & 185K \\
    DART & graph-to-text & open & 70,524 & 34.8 & 44K & 2.5M & 19.3 & 45K & 1.5M \\
    WebNLG & graph-to-text & open & 38,872 & 30.4 & 7K & 1.2M & 20.1 & 19K & 905K \\ \bottomrule
    \end{tabular}}
    \caption{Summary of key statistics, D2T types, and domains for the five incorporated datasets. The table presents average length (number of tokens in text), unique tokens, and total tokens for both sources (linearized to text) and references.}
    \label{tab:dataset_statistics}
\end{table}

\paragraph{DART dataset.}
The DART dataset~\citep{nan2021dart} is integral for open-domain structured D2T generation, similar to the WikiTableText dataset. 
It pertains specifically to the graph-to-text D2T task, aiming to generate text from knowledge graph triplets. 
Each instance in the dataset covers a variety of domains, with inputs consisting of semantic triplet sets accompanied by detailed sentence descriptions. 
These descriptions are curated from a multitude of datasets. On average, the reference text sets contain approximately 19.3 tokens each.

\paragraph{WebNLG dataset.}
The underlying task of this dataset involves generating text from a series of knowledge graph entries, also known as RDF (Resource Description Format)-to-text generation, a type of graph-to-text generation~\citep{gardent2017webnlg}. 
The dataset focuses on generating text from knowledge graph triplets sourced mainly from DBPedia across six domains~\citep{perezbeltrachini2017analysing,gardent2017webnlg}. 
It's gathered via crowdsourcing, ensuring diverse and validated outputs.
Notably, it exhibits greater lexical and semantic diversity compared to similar datasets.

\subsection{Experimental Settings}
In all our experiments in this study, we fined-tune LLMs for D2T task. 
We fine-tune each LLM on all five datasets for 3 epochs.
Because of smaller sizes, T5 and BART models underwent full fine-tuning with a learning rate of 1e-05.
Both T5 and BART were loaded with \texttt{torch.bfloat16} precision.
However, due to the large number of parameters in BLOOM, OPT, and Llama 2, we employed the QLoRA (Quantized version of Low Rank Adapter)~\citep{dettmers2023qlora} technique as a parameter-efficient fine-tuning approach, utilizing a learning rate of 1e-04. 
All these three larger LLM families' models were loaded with 4-bit precision.
The peft library are used for implementing QLoRA in our experiments~\citep{mangrulkar2022peft}.
As previously mentioned, all 12 LLMs and the five D2T datasets were sourced from HuggingFace~\citep{wolf2020transformers}.
We maintained a consistent text length of 256 for all generations, encompassing both source data and generated text. 
For optimization purposes, we utilized the popular AdamW optimizer~\citep{loshchilov2019decoupled}, which is primarily based on the Adam optimizer~\citep{kingma2015adam} with $L^2$-norm over the weight space.
In QLoRA set-ups, we continue AdamW optimizer's operation at \texttt{torch.bfloat16} precision.
Decoding strategies play a crucial role in text generation.
We compared all generated text using the beam-search decoding strategy with a beam size of 5.
All experiments were conducted on a local system equipped with an NVIDIA RTX A6000 (48 GB) graphics card to ensure adequate GPU access.

\paragraph{Statistical Significant Testing.}
We aim to determine the statistical significance of the performance evaluations in all our comparative analyses across the three key qualities of D2T (\textit{readability}, \textit{informativeness}, and \textit{faithfulness}).
To achieve this, we conduct Welch's t-test~\citep{dror2018hitchhikers} comparing the results of each LLM to the best-performing LLM within the same family. 
We apply a significance level of $p < 0.05$ for these tests, with each Welch's t-test based on samples size of 6.

\section{Performance Evaluations and Results}
\label{sec:evaluation_results}
In this section, we analyze the performance of LLMs in D2T using popularly used evaluation metrics across three aspects: \textit{Readability}, \textit{Informativeness}, and \textit{Faithfulness}.

\subsection{Readability}
The \textit{readability}~\citep{reiter1997building,belz2006comparing,chen2020logic2text} of a D2T model mainly considers two factors: fluency and coherence. 
Fluency determines whether the generated text is grammatically correct and easy to read, enhancing its naturalness in terms of reading. 
Coherence assesses whether the information, topics, and facts in the generated text are well-ordered and chronologically expressed.
To measure \textit{readability}, we incorporate two automatic evaluation metrics: \textsc{Bleu} and \textsc{Meteor}. We explain these two metrics briefly below.

\begin{table}[ht]
    \centering
    \resizebox{0.8\textwidth}{!}{
    \begin{tabular}{@{}cccccccc@{}}
        \toprule
        family & model & \begin{tabular}[c]{@{}c@{}}size (number of\\ parameters in billion)\end{tabular} & E2E & ViGGo & WikiTableText & DART & WebNLG \\ \midrule
        \multirow{2}{*}{BART} & BART-base & 0.1 & 0.399 & 0.281 & \textbf{0.421} & \textbf{0.423} & 0.481 \\
         & BART-large & 0.4 & \textbf{0.403} & \textbf{0.283} & 0.419 & 0.413 & \textbf{0.503} \\ \midrule
        \multirow{2}{*}{T5} & T5-base & 0.2 & 0.398 & 0.268 & 0.408 & 0.461 & 0.527 \\
         & T5-large & 0.7 & \textbf{0.411} & \textbf{0.302} & \textbf{0.431} & \textbf{0.479} & \textbf{0.546} \\ \midrule
        \multirow{3}{*}{OPT} & OPT-2.7B & 2.7 & 0.350 & 0.262 & 0.421 & 0.441 & 0.521 \\
         & OPT-6.7B & 6.7 & \textbf{0.369} & \textbf{0.269} & \textbf{0.426} & 0.448 & 0.538 \\
         & OPT-13B & 13.0 & 0.347 & 0.269 & 0.412 & \textbf{0.463} & \textbf{0.549} \\ \midrule
        \multirow{3}{*}{BLOOM} & BLOOM-1.1B & 1.1 & 0.374 & 0.255 & 0.411 & 0.437 & 0.491 \\
         & BLOOM-3B & 3.0 & \textbf{0.380} & 0.260 & 0.396 & \textbf{0.446} & 0.520 \\
         & BLOOM-7B & 7.0 & 0.379 & \textbf{0.274} & \textbf{0.423} & 0.444 & \textbf{0.530} \\ \midrule
        \multirow{2}{*}{Llama 2} & Llama2-7B & 7.0 & \textbf{0.419} & 0.248 & 0.436 & 0.494 & 0.532 \\
         & Llama2-13B & 13.0 & 0.408 & \textbf{0.288} & \textbf{0.451} & \textbf{0.51} & \textbf{0.563} \\ \bottomrule
    \end{tabular}}
    \caption{Analysis of \textit{readability} across 12 LLMs from five LLM families (BERT, T5, BLOOM, OPT, and Llama 2) on all five D2T datasets, measured using the \textsc{Bleu} metric. The best scoring results in terms of model size for each family are highlighted in bold. Overall, the \textsc{Bleu} score consistently tends to increase with larger model sizes, except for a few exceptions. All results are statistically significant at the $p<0.05$ level.}
    \label{tab:bleu}
\end{table}

\begin{itemize}
    \item \underline{\textsc{Bleu}}. 
    \textsc{Bleu}~\citep{papineni2002bleu} evaluates the \textit{readability} between generated and reference texts using n-gram matching. 
    It measures both precision and recall based on n-gram overlap. 
    Despite facing criticism, \textsc{Bleu} remains widely used for assessing textual fluency.
    Essentially, \textsc{Bleu} computes the geometric mean of n-gram precision of generated text with respect to the reference.
    Additionally, a brevity penalty term is incorporated into \textsc{Bleu} to address length discrepancies between the generated and reference texts.
    We can represent \textsc{Bleu} as the following equation:
    
    \begin{align*}
        \textsc{Bleu} = \beta \cdot \text{exp}\left(\sum_n \text{log}(w_n \cdot \text{Precision}_n)\right)
    \end{align*}

    where $\text{Precision}_n$ represents the precision based on $n$-grams, $w_n$ denotes the weight associated with the corresponding precision, and $\beta$ represents the brevity penalty (defines through equation~\ref{eq:bleu_brevity}). 

    \begin{align}
        \beta =
        \begin{cases}
            1 & \text{if $|g| > |r|$}\\
            \text{exp} \left(1-\frac{\vert r|}{|g|}\right) & \text{otherwise}
        \end{cases} \label{eq:bleu_brevity}
    \end{align}
    
    Where $|r|$ and $|g|$ denote the lengths of the reference ($r$) and generated ($g$) text, respectively.
    In most cases, $n$ is set to 4, which is also known as \textsc{Bleu}-4.
    In our evaluation also we employed \textsc{Bleu}-4.
    We used SacreBLEU~\citep{post2018call} implementation to obtain our blue results.

    \item \underline{\textsc{Meteor}}. \textsc{Meteor}~\citep{banerjee2005meteor}, similar to \textsc{Bleu}, relies on n-gram string matching and finds predominant use in machine translation tasks.
    While \textsc{Bleu} focuses solely on n-gram precision, it overlooks the recall component.
    \textsc{Meteor} addresses this by leveraging unigram-based string overlaps to compute both recall and precision.
    Ultimately, \textsc{Meteor} distinguishes itself as a weighted variant of the F1-measure, as expressed by the following equation:

    \begin{align*}
        \textsc{Meteo}_\text{F1-measure} = \frac{10 \cdot \text{Precison} \cdot \text{Recall}}{\text{Recall} + 9\text{Precison}}
    \end{align*}
    Where, $\text{Precision}$ and $\text{Recall}$ denote precision and recall based on unigram-based matching.
    To account for higher-order $n$-grams, \textsc{Meteor} applies a penalty factor to the F1-measure.
    The penalty factor is calculated as the ratio of the number of chunks concatenated through matched unigrams to the number of matched unigrams (equation~\ref{eq:meteor_penalty}).
    \begin{align}
        \text{penalty} = 0.5 \times \frac{\text{number of chunks}}{\text{number of matched unigrams}} \label{eq:meteor_penalty}
    \end{align}

    Finally, \textsc{Meteor} score is calculated through,
    \begin{align*}
        \textsc{Meteor} = (1-\text{penalty}) \cdot \textsc{Meteo}_\text{F1-measure}
    \end{align*}
    
    In addition to the above formulation, unlike \textsc{Bleu}, \textsc{Meteor} integrates synonym matching (via WordNet synsets) and word stemming (via Porter's stemmer) features to capture textual similarity between candidates and references.

    \begin{table}[ht]
        \centering
        \resizebox{0.8\textwidth}{!}{
        \begin{tabular}{@{}cccccccc@{}}
            \toprule
            family & model & \begin{tabular}[c]{@{}c@{}}size (number of\\ parameters in billion)\end{tabular} & E2E & ViGGo & WikiTableText & DART & WebNLG \\ \midrule
            \multirow{2}{*}{BART} & BART-base & 0.1 & 0.368 & 0.315 & 0.367 & 0.365 & 0.377 \\
             & BART-large & 0.4 & \textbf{0.371} & \textbf{0.321} & \textbf{0.367} & \textbf{0.381} & \textbf{0.408} \\ \midrule
            \multirow{2}{*}{T5} & T5-base & 0.2 & 0.369 & 0.312 & 0.361 & 0.390 & 0.410 \\
             & T5-large & 0.7 & \textbf{0.377} & \textbf{0.324} & \textbf{0.370} & \textbf{0.400} & \textbf{0.420} \\ \midrule
            \multirow{3}{*}{OPT} & OPT-2.7B & 2.7 & 0.355 & 0.307 & 0.366 & 0.382 & 0.403 \\
             & OPT-6.7B & 6.7 & \textbf{0.366} & 0.305 & \textbf{0.371} & 0.387 & 0.406 \\
             & OPT-13B & 13.0 & 0.355 & \textbf{0.308} & 0.361 & \textbf{0.391} & \textbf{0.411} \\ \midrule
            \multirow{3}{*}{BLOOM} & BLOOM-1.1B & 1.1 & 0.369 & 0.304 & 0.367 & 0.379 & 0.386 \\
             & BLOOM-3B & 3.0 & \textbf{0.371} & 0.305 & 0.356 & \textbf{0.386} & 0.403 \\
             & BLOOM-7B & 7.0 & 0.369 & \textbf{0.314} & \textbf{0.375} & 0.385 & \textbf{0.412} \\ \midrule
            \multirow{2}{*}{Llama 2} & Llama2-7B & 7.0 & \textbf{0.376} & 0.307 & 0.375 & 0.403 & 0.412 \\
             & Llama2-13B & 13.0 & 0.37 & \textbf{0.318} & \textbf{0.38} & \textbf{0.407} & \textbf{0.419} \\ \bottomrule
        \end{tabular}}
        \caption{Analysis of the \textit{readability} of 12 LLMs from five LLM families (BERT, T5, BLOOM, OPT, and Llama 2) across all five D2T datasets, evaluated using the \textsc{Meteor} metric. The best-performing results in terms of model size for each family are highlighted in bold. Similar to the \textsc{Bleu} results, the \textsc{Meteor} score tends to increase with larger model sizes, with a few exceptions. All results are statistically significant at the $p<0.05$ level.}
        \label{tab:meteor}
    \end{table}
\end{itemize}

\paragraph{Takeaways.}
From our empirical results on \textit{readability} based on Table~\ref{tab:bleu} and Table~\ref{tab:meteor}, we can derive insights to address the \textit{readability} part of first question posed in Section~\ref{sec:aim_motivation} .

\begin{itemize}
    \item [] \textbf{Question:} What are the impacts of model size within a family of fine-tuned LLMs on the performance of data-to-text (D2T) tasks, in terms of the \underline{\textit{readability}} \textcolor{gray}{, \textit{informativeness}, and \textit{faithfulness}}?
    \item [] \textbf{Answer:} Undoubtedly, the parameter count of LLMs significantly influences the \textit{readability} of D2T tasks. 
    Across all three categories of D2T tasks, it becomes apparent from Table~\ref{tab:bleu} and Table~\ref{tab:meteor} that augmenting the model parameters substantially enhances the \textit{readability} of D2T tasks, with only minor discrepancies observed.
    Nearly all LLMs demonstrate superior performance compared to their lower-sized counterparts within the same LLM family. 
    Consequently, the \textsc{Bleu} and \textsc{Meteor} scores consistently indicate improved \textit{readability} across all LLMs within their respective families.
    If we consider the sum of all results across the five datasets, it is evident that in a family of large language models (LLMs), the \textit{readability} scores consistently improve with an increase in the number of parameters, i.e., model size. As both Table~\ref{tab:bleu} and Table~\ref{tab:meteor} show statistically significant results at the $p < 0.05$ level, thereby reinforcing our findings.
\end{itemize}

\subsection{Informativeness}
Informativeness of D2T models concerns the content similarity between reference text and the generated text from the model~\citep{shen2019pragmatically,erdem2022neural,belz2020disentangling}.
A highly informative D2T model indicates its ability to generate genuinely helpful information.
Semantic similarity between reference and generated text contributes to content similarity.
To measure informativeness based on content similarity, we incorporate two contextual similarity-based metrics---\textsc{BertScore} and \textsc{MoverScore}.
It's important to note that, similar to reference text, source text also contains valuable information, and in the presence of source-reference divergence, it becomes crucial to assess the similarity between source and generated text, similar to reference text.
For measuring informativeness in the presence of source-reference divergence, we use the \textsc{Parent} metric, which considers both reference and source text to compute content similarity with respect to the generated text.

\begin{itemize}
    \item \underline{\textsc{BertScore}}. \textsc{BertScore}~\citep{zhang2020bertscore} leverages contextual representations to assess text similarity.
    Unlike models based on n-gram matching, which may struggle to capture contextual nuances, \textsc{BertScore} utilizes contextual representations derived from the BERT model.
    BERT is trained using masked language modeling.
    Contextual representations of words within a text can be extracted from either the hidden layers or output layers of BERT.
    \textsc{BertScore} utilizes these contextual representations and employs the cosine similarity function to estimate the similarity between two contextual representations.
    It calculates an F1-score based on contextual matching between the generated text ($g$) and the reference text ($r$).
    Lets, assumed $g$ contains $m$ words---$g_1g_2 \dots g_m$, and $r$ contains $n$ words---$r_1r_2 \dots r_n$,
    Contextual representation of BERT of $g$ is, 
    \begin{align*}
        \text{BERT}(g) = h[g_1] h[g_2] \dots h[g_m]
    \end{align*}

    Likewise we can obtained $\text{BERT}(r)$ as $h[r_1] h[r_2] \dots h[r_n]$. 
    To compute the recall of \textsc{BertScore}, we perform the following steps:
    \begin{align*}
        \text{Recall}_{\textsc{BertScore}} = \frac{1}{|r|} \sum_{r_i \in r} \underset{g_j \in g}{\text{max}} (h[r_i]^{\mathsf{T}} h[g_j])
    \end{align*}

    Similarly, the precision of \textsc{BertScore} is measured as follows:
    \begin{align*}
        \text{Precision}_{\textsc{BertScore}} = \frac{1}{|g|} \sum_{g_j \in g} \underset{r_i \in r}{\text{max}} (h[r_i]^{\mathsf{T}} h[g_j])
    \end{align*}

    Finally, \textsc{BertScore} is obtained by calculating the F1-measure of $\text{Precision}_{\textsc{BertScore}}$ and $\text{Recall}_{\textsc{BertScore}}$.
    Widely regarded as a superior metric for semantic similarity, BERTScore outperforms several other metrics.

    \begin{table}[ht]
        \centering
        \resizebox{0.8\textwidth}{!}{
        \begin{tabular}{@{}cccccccc@{}}
            \toprule
            family & model & \begin{tabular}[c]{@{}c@{}}size (number of\\ parameters in billion)\end{tabular} & E2E & ViGGo & WikiTableText & DART & WebNLG \\ \midrule
            \multirow{2}{*}{BART} & BART-base & 0.1 & 0.918 & 0.889 & \textbf{0.899} & 0.908 & 0.919 \\
             & BART-large & 0.4 & \textbf{0.918} & \textbf{0.890} & 0.896 & \textbf{0.914} & \textbf{0.927} \\ \midrule
            \multirow{2}{*}{T5} & T5-base & 0.2 & 0.917 & 0.884 & 0.895 & 0.925 & 0.932 \\
             & T5-large & 0.7 & \textbf{0.919} & \textbf{0.895} & \textbf{0.899} & \textbf{0.927} & \textbf{0.936} \\ \midrule
            \multirow{3}{*}{OPT} & OPT-2.7B & 2.7 & 0.908 & 0.879 & 0.896 & 0.919 & 0.927 \\
             & OPT-6.7B & 6.7 & \textbf{0.910} & 0.874 & \textbf{0.898} & 0.919 & 0.927 \\
             & OPT-13B & 13.0 & 0.908 & \textbf{0.879} & 0.897 & \textbf{0.920} & \textbf{0.928} \\ \midrule
            \multirow{3}{*}{BLOOM} & BLOOM-1.1B & 1.1 & 0.911 & 0.884 & 0.895 & 0.915 & 0.917 \\
             & BLOOM-3B & 3.0 & \textbf{0.912} & 0.883 & 0.887 & 0.916 & 0.925 \\
             & BLOOM-7B & 7.0 & 0.911 & \textbf{0.886} & \textbf{0.896} & \textbf{0.917} & \textbf{0.928} \\ \midrule
            \multirow{2}{*}{Llama 2} & Llama2-7B & 7.0 & 0.918 & 0.882 & 0.897 & 0.931 & 0.934 \\
             & Llama2-13B & 13.0 & \textbf{0.918} & \textbf{0.889} & \textbf{0.900} & \textbf{0.931} & \textbf{0.938} \\ \bottomrule
        \end{tabular}}
        \caption{Comparative evaluation of \textit{Informativeness} for 12 LLMs from five families (BERT, T5, BLOOM, OPT, and Llama 2) using the \textsc{BertScore} metric on five D2T datasets. Bold highlights indicate the best performing model sizes within each family. Generally, \textsc{BertScore} increases with larger model sizes, with minor exceptions. All results are statistically significant at $p < 0.05$.}
        \label{tab:bertscore}
    \end{table}

    \item \underline{\textsc{MoverScore}}. 
    Similar to \textsc{BertScore}, \textsc{MoverScore}~\citep{zhao2019moverscore} prioritizes contextual similarity between texts.
    It quantifies text embeddings derived from the contextual representation of all underlying words using power law.
    To compute the semantic distance between two texts, it utilizes Word Mover's Distance (WMD) with the underlying $n$-grams of the texts (equation~\ref{eq:moverscore_wmd}).

    \begin{align}
        \text{WMD} (g, r) = \underset{F \in \mathbb{R}^{|g| \times |r|}}{\text{min}} \sum C \odot F \label{eq:moverscore_wmd}
    \end{align}

    Where $C$ and $F$ denote cost matrix and flow matrix. 
    In WMD, each entry of the cost matrix is determined by the contextual representation of the n-grams obtained after applying the power law.
    The flow metrics are obtained by determining the weight of each n-gram through inverse document frequency.

    \begin{table}[ht]
        \centering
        \resizebox{0.8\textwidth}{!}{
        \begin{tabular}{@{}cccccccc@{}}
            \toprule
            family & model & \begin{tabular}[c]{@{}c@{}}size (number of\\ parameters in billion)\end{tabular} & E2E & ViGGo & WikiTableText & DART & WebNLG \\ \midrule
            \multirow{2}{*}{BART} & BART-base & 0.1 & 0.667 & 0.658 & \textbf{0.688} & 0.666 & 0.673 \\
             & BART-large & 0.4 & \textbf{0.667} & \textbf{0.664} & 0.683 & \textbf{0.671} & \textbf{0.684} \\ \midrule
            \multirow{2}{*}{T5} & T5-base & 0. & 0.666 & 0.650 & 0.679 & 0.688 & 0.685 \\
             & T5-large & 0.7 & \textbf{0.672} & \textbf{0.669} & \textbf{0.683} & \textbf{0.692} & \textbf{0.691} \\ \midrule
            \multirow{3}{*}{OPT} & OPT-2.7B & 2.7 & 0.655 & 0.645 & 0.682 & 0.673 & 0..676 \\
             & OPT-6.7B & 6.7 & \textbf{0.659} & 0.639 & \textbf{0.687} & 0.673 & 0.678 \\
             & OPT-13B & 13.0 & 0.654 & \textbf{0.646} & 0.684 & \textbf{0.676} & \textbf{0.679} \\ \midrule
            \multirow{3}{*}{BLOOM} & BLOOM-1.1B & 1.1 & 0.658 & 0.649 & 0.676 & 0.669 & 0.663 \\
             & BLOOM-3B & 3.0 & 0.658 & 0.651 & 0.670 & 0.670 & 0.673 \\
             & BLOOM-7B & 7.0 & \textbf{0.658} & \textbf{0.655} & \textbf{0.678} & \textbf{0.671} & \textbf{0.679} \\ \midrule
            \multirow{2}{*}{Llama 2} & Llama2-7B & 7.0 & \textbf{0.671} & 0.651 & 0.686 & 0.691 & 0.689 \\
             & Llama2-13B & 13.0 & 0.668 & \textbf{0.66} & \textbf{0.69} & \textbf{0.692} & \textbf{0.691} \\ \bottomrule
        \end{tabular}}
        \caption{Assessment of \textit{Informativeness} across 12 LLMs from five families (BERT, T5, BLOOM, OPT, and Llama 2) using the \textsc{MoverScore} metric on five D2T datasets. The highest-performing model sizes within each family are highlighted in bold. Consistent with the \textsc{BertScore} metric, \textsc{MoverScore} generally increases with model size, reflecting improved \textit{informativeness}. All results are statistically significant at $p < 0.05$.}
        \label{tab:moverscore}
    \end{table}

    \item \underline{\textsc{Parent}}. Compared to \textsc{BertScore} and \textsc{MoverScore}, \textsc{Parent}~\citep{dhingra2019handling} assesses the generated text using both reference and source texts.
    It is particularly valuable in scenarios with source-reference divergence and to verify if the generated text aligns with the source data.
    \textsc{Parent} correlates strongly with human judgments, as it considers both reference and source data.
    It calculates an F1-measure based on recall and precision, computed using lexical entailment techniques such as word overlap and co-occurrence.
    Let's assume that the lexical entailment of $g_i$ (where $g_i$ is an $n$-gram of $g$, i.e., $g_i \in \text{n-gram}(g)$) with respect to $s$ and $r$ is represented through ${E}_n(g_i, s)$ and ${E}_n(g_i, r)$.
    Then, the entailed precision of the generated text ($g$) with respect to both the source data ($s$) and the reference text ($r$) is defined as follows:
    \begin{align*}
        \text{Precision}_{{E}_n} (g, s, r) \propto\ \frac{\sum\limits_{g_i \in \text{n-gram}(g)}{E}_n(g_i, r) + (1-{E}_n(g_i, r)) {E}_n(g_i, s)}{|g_i \in \text{n-gram}(g)|}
    \end{align*}
    
    Likewise, the entailed recall of $ g $ with respect to both $ s $ and $ r $ is calculated as follows:
    \begin{align*}
        \text{Recall}_{{E}_n} (g, s, r) = \text{Recall}_{{E}_n} (g, s)^{\lambda} \cdot \text{Recall}_{{E}_n}(g, r)^{1-\lambda}
    \end{align*}
    where $\lambda$ is the weight factor of the two recalls. $\text{Recall}_{{E}_n}(G, R)$ is calculated through lexical entailment, where $\text{Recall}_{{E}_n} (G, S)$ is calculated through longest common sub-sequence matching, as source ($s$) often is in semi-structured data.
    Finally, \textsc{Parent} is measure through F1-measure of $\text{Precision}_{{E}_n}$ and $\text{Recall}_{{E}_n}$.

    \begin{table}[ht]
        \centering
        \resizebox{0.8\textwidth}{!}{
        \begin{tabular}{@{}cccccccc@{}}
            \toprule
            family & model & \begin{tabular}[c]{@{}c@{}}size (number of\\ parameters in billion)\end{tabular} & E2E & ViGGo & WikiTableText & DART & WebNLG \\ \midrule
            \multirow{2}{*}{BART} & BART-base & 0.1 & \textbf{0.613} & \textbf{0.420} & \textbf{0.526} & 0.538 & 0.567 \\
             & BART-large & 0.4 & 0.600 & 0.413 & 0.514 & \textbf{0.563} & \textbf{0.586} \\ \midrule
            \multirow{2}{*}{T5} & T5-base & 0.2 & 0.599 & 0.430 & 0.533 & 0.614 & 0.618 \\
             & T5-large & 0.738 & \textbf{0.603} & \textbf{0.436} & \textbf{0.547} & \textbf{0.622} & \textbf{0.644} \\ \midrule
            \multirow{3}{*}{OPT} & OPT-2.7B & 2.7 & 0.528 & 0.395 & \textbf{0.510} & 0.557 & 0.589 \\
             & OPT-6.7B & 6.7 & \textbf{0.549} & \textbf{0.401} & 0.507 & 0.559 & 0.600 \\
             & OPT-13B & 13.0 & 0.529 & 0.399 & 0.499 & \textbf{0.565} & \textbf{0.612} \\ \midrule
            \multirow{3}{*}{BLOOM} & BLOOM-1.1B & 1.1 & 0.565 & \textbf{0.405} & 0.502 & \textbf{0.561} & 0.566 \\
             & BLOOM-3B & 3.0 & \textbf{0.569} & 0.400 & 0.476 & 0.556 & 0.585 \\
             & BLOOM-7B & 7.0 & 0.565 & 0.404 & \textbf{0.504} & 0.557 & \textbf{0.594} \\ \midrule
            \multirow{2}{*}{Llama 2} & Llama2-7B & 7.0 & \textbf{0.602} & 0.390 & 0.531 & 0.621 & 0.621 \\
             & Llama2-13B & 13.0 & 0.599 & \textbf{0.418} & \textbf{0.531} & \textbf{0.622} & \textbf{0.640} \\ \bottomrule
        \end{tabular}}
        \caption{Assessment of \textit{informativeness} across 12 LLMs from five LLM families on all five D2T datasets, using the \textsc{Parent} metric. The best-performing model sizes within each family are highlighted in bold. Unlike \textsc{BertScore} and \textsc{MoverScore}, \textsc{Parent} shows a mixed response, with inconsistent behavior observed in some LLM families such as BART, OPT, and BLOOM. However, in most cases, \textsc{Parent} exhibits a positive correlation with increasing model size. All results are statistically significant at $p < 0.05$.}
        \label{tab:parent}
    \end{table}
\end{itemize}

\paragraph{Takeaways.}
Except some negligible exceptions, the empirical findings from Tables~\ref{tab:bertscore}, \ref{tab:moverscore}, and \ref{tab:parent} consistently validate the relationship between model parameters and the informativeness of D2T models. 
As the model parameters increase, there is a clear and noticeable improvement in the LLM's informativeness. 
With this comprehensive understanding, we are now ready to address \textit{informativeness} part the first question posed in Section~\ref{sec:aim_motivation}.

\begin{itemize}
    \item [] \textbf{Question:} What are the impacts of model size within a family of fine-tuned LLMs on the performance of data-to-text (D2T) tasks, in terms of the \textcolor{gray}{\textit{readability},} \underline{\textit{informativeness}}\textcolor{gray}{, and \textit{faithfulness}}?
    \item [] \textbf{Answer:} Both Table~\ref{tab:bertscore} and Table~\ref{tab:moverscore} present compelling evidence supporting the notion that increasing model size enhances the \textit{informativeness} of LLMs in the same family, particularly concerning the alignment between reference and generated text.
    Across all three primary D2T tasks, a consistent improvement in \textit{informativeness} is observed with the size of LLMs.
    While isolated cases of decreased \textit{informativeness} with increasing LLM parameters are noted, such as with BART family (in the case of the WikiTableText dataset using both \textsc{BertScore} and \textsc{MoverScore}), and Llama 2 family (in the case of the E2E dataset using \textsc{MoverScore}), these instances exhibit minimal degradation.
    Consequently, they do not significantly challenge our conclusion that increasing number of parameters enhances the \textit{informativeness} of LLMs inside of a LLM family for D2T tasks.
    Furthermore, Table~\ref{tab:parent} also indicates that increasing size often leads to enhanced \textit{informativeness} between source and generated text. 
    However, compared to \textsc{BertScore} and \textsc{MoverScore}, discrepancies are more pronounced. 
    Some LLMs from BART, OPT, and BLOOM families exhibit instances where increasing parameters may degrade \textsc{Parent} scores (see Table~\ref{tab:parent}).
    Therefore, it is apparent that enlarging model size may occasionally diminish the \textit{informativeness} inside of a LLM family when \textit{informativeness} evaluated between source and generated text.
\end{itemize}

\subsection{Faithfulness}
\begin{figure}[ht]
    \centering
    \resizebox{0.6\textwidth}{!}{
    \includegraphics{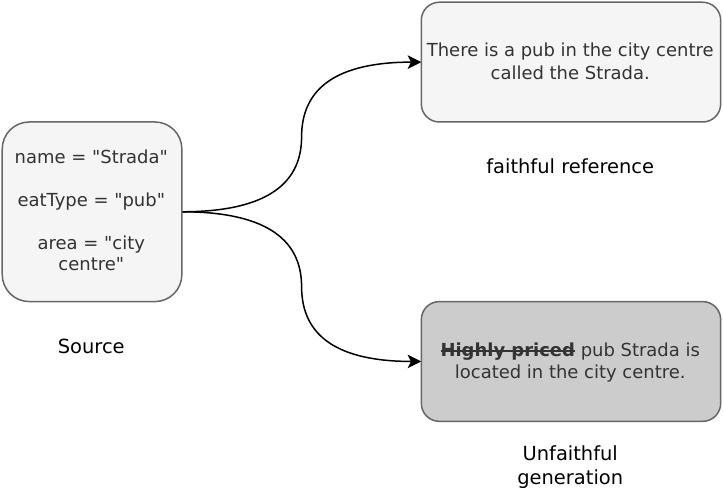}}
    \caption{An illustration of unfaithful generation on the E2E dataset. The generated text contains the phrase ``\textbf{Highly price}", which is irrelevant with respect to the information provided in the source data.}
    \label{fig:unfaithful}
\end{figure}

\textit{Faithfulness} is a crucial aspect of D2T model~\citep{wang2020towards,ji2023survey}, primarily assessing factual consistency or accuracy of the generated text compared to the provided source data.
It can be characterized by two cases: i) the model generating incorrect facts compared to the given source data, and ii) the model generating irrelevant facts that do not appear in the source data.
Figure~\ref{fig:unfaithful} illustrates an instance of unfaithful generation on the E2E dataset.
To quantify \textit{faithfulness} of a D2T model, we employ the \textsc{BartScore} metric, which utilizes the likelihood of a BART model for estimating \textit{faithfulness}.

\begin{itemize}
    \item \underline{\textsc{BartScore}}. \textsc{BartScore}~\citep{yuan2021bartscore} evaluates the \textit{faithfulness} between two texts by estimating the likelihood of text generation based on a sequence-to-sequence model like BART.
    Its premise is that if two pieces of text are factually equivalent, the likelihood of generating one from the other will be higher.
    \textit{Faithfulness} in \textsc{BartScore} is determined by the probability (equation~\ref{eq:bartscore}) of transforming from the source text ($s$) to the generated text ($g$).

    \begin{align}
        \text{Faithfulness} (s, g) & = Pr(\text{BART}(S \rightarrow G)) \nonumber\\
         & = \left(\prod_{g_i \in g} Pr \left(g_i \leftarrow \text{BART}(s, g_1, g_2, \dots, g_{i-1})\right)\right)^{\frac{1}{|g|}} \label{eq:bartscore}
    \end{align}
    
    \textsc{BartScore} has exhibited superior performance compared to several earlier evaluation metrics. To calculate \textsc{BartScore}, we employ a pre-trained BARTScore model trained on the PARABANK dataset~\citep{yuan2021bartscore}.
    It's essential to note that when dealing with source ($s$) as semi-structured data, before computing the \textsc{BartScore} of the generated text $g$ with respect to $s$, we must linearized $s$ through text.

    \begin{table}[ht]
        \centering
        \resizebox{0.8\textwidth}{!}{
        \begin{tabular}{@{}cccccccc@{}}
            \toprule
            family & model & \begin{tabular}[c]{@{}c@{}}size (number of\\ parameters in billion)\end{tabular} & E2E & ViGGo & WikiTableText & DART & WebNLG \\ \midrule
            \multirow{2}{*}{BART} & BART-base & 0.1 & 0.088 & \textbf{0.047} & \textbf{0.066} & 0.061 & \textbf{0.097} \\
             & BART-large & 0.4 & \textbf{0.092} & 0.046 & 0.066 & \textbf{0.065} & 0.094 \\ \midrule
            \multirow{2}{*}{T5} & T5-base & 0.2 & 0.091 & \textbf{0.052} & \textbf{0.072} & 0.067 & 0.093 \\
             & T5-large & 0.7 & \textbf{0.093} & 0.048 & 0.072 & \textbf{0.068} & \textbf{0.095} \\ \midrule
            \multirow{3}{*}{OPT} & OPT-2.7B & 2.7 & \textbf{0.057} & 0.045 & \textbf{0.068} & \textbf{0.063} & \textbf{0.089} \\
             & OPT-6.7B & 6.7 & 0.053 & \textbf{0.046} & 0.066 & 0.062 & 0.087 \\
             & OPT-13B & 13.0 & 0.050 & 0.043 & 0.065 & 0.063 & 0.087 \\ \midrule
            \multirow{3}{*}{BLOOM} & BLOOM-1.1B & 1.1 & \textbf{0.089} & \textbf{0.042} & \textbf{0.057} & \textbf{0.064} & \textbf{0.090} \\
             & BLOOM-3B & 3.0 & 0.083 & 0.041 & 0.054 & 0.064 & 0.089 \\
             & BLOOM-7B & 7.0 & 0.086 & 0.040 & 0.056 & 0.063 & 0.088 \\ \midrule
            \multirow{2}{*}{Llama 2} & Llama2-7B & 7.0 & \textbf{0.094} & \textbf{0.05} & \textbf{0.062} & 0.073 & \textbf{0.106} \\
             & Llama2-13B & 13.0 & 0.089 & 0.048 & 0.061 & \textbf{0.074} & 0.105 \\ \bottomrule
        \end{tabular}
        }
        \caption{Assessment of \textit{faithfulness} across 12 LLMs from five families (BERT, T5, BLOOM, OPT, and Llama 2) on the five D2T datasets, using probabilities from the \textsc{BartScore} metric (noting that some studies report \textsc{BartScore} using logits). The best-performing model sizes within each family are highlighted in bold. Unlike \textit{readability} and \textit{informativeness}, \textsc{BartScore} shows an inverse trend for \textit{faithfulness}. In most LLM families, increasing model size correlates with a decrease in \textsc{BartScore}, indicating reduced \textit{faithfulness} in larger models. All results are statistically significant at $p < 0.05$.}
        \label{tab:bartscore}
    \end{table}
\end{itemize}

\paragraph{Takeaways.}
Upon scrutinizing the empirical findings presented in Table~\ref{tab:bartscore}, we are now equipped to address \textit{faithfulness} part of the first question posed in Section~\ref{sec:aim_motivation} concerning the \textit{faithfulness} of D2T models.
\begin{itemize}
    \item [] \textbf{Question:} What are the impacts of model size within a family of fine-tuned LLMs on the performance of data-to-text (D2T) tasks, in terms of the \textcolor{gray}{\textit{readability}, \textit{informativeness}, and} \underline{\textit{faithfulness}}?
    \item [] \textbf{Answer:} No, an increase in the number of parameters (i.e., model size) will not necessarily improve the \textit{faithfulness} of LLMs inside of a LLM family for D2T tasks.
    In fact, increasing parameters can degrade the performance of LLMs in terms of \textit{faithfulness}.
    Table~\ref{tab:bartscore} clearly suggests that for LLMs from BLOOM, OPT, and Llama 2 families, increasing their parameters adversely affects their \textit{faithfulness} aspect for all D2T datasets.
    Although we have seen a very small number of discrepancies (five cases with a very low margin) in T5 and BART families (for the E2E and DART datasets) and Llama 2 family (for the DART dataset), they still indicate that \textit{faithfulness} often degrades with increasing parameters on other datasets.
    Hence, we can certainly claim that increasing the number of parameters will not be helpful for enhancing the \textit{faithfulness} of LLMs toward D2T.
\end{itemize}

\subsection{Comparative Performance Analysis across LLM Families}
Thus far, our focus has been primarily on analyzing the performance of LLMs within individual LLM families.
Now, our attention shifts towards the comparative performance analysis of LLMs across different LLM families.
Considering all six metrics' tables, Table ~\ref{tab:bleu}, ~\ref{tab:meteor}, ~\ref{tab:bertscore}, ~\ref{tab:moverscore}, ~\ref{tab:parent}, and ~\ref{tab:bartscore}, we can effectively address the second question raised in Section~\ref{sec:aim_motivation}.

\begin{itemize}
    \item [] \textbf{Question:} Do larger LLM families (such as OPT, BLOOM, Llama 2, etc.) convincingly outperform smaller LLM families (such as BART, T5, etc.) in terms of D2T task performance?
    \item [] \textbf{Answer:} From the results of all six automatic metrics, there is no significant evidence to conclusively state that larger LLM families (from BLOOM, OPT, and Llama 2 families) outperform smaller LLM families (from BART and T5).
    While Llama 2 family shows significant improvement over T5 and BART families in terms of \textit{readability} and \textit{informativeness}, it falls short compared to \textsf{T5-large} model of T5 family in terms of \textit{faithfulness}.
    On the other hand, both OPTs and BLOOMs families perform lower than the T5 family across all three quality aspects.
    Consequently, it can be inferred that LLM from smaller model sized family, such as \textsf{T5-large} model, can surprisingly perform quite well in D2T tasks.
    Moreover, their main advantage lies in their lower computational cost compared to larger LLM families like Llama 2.
    Especially for faithful D2T tasks~\citep{wang2020towards,ji2023survey}, it is preferable to utilize such smaller LLM families.
\end{itemize}

\subsection{Discussion}
From the empirical results of all six automatic metrics on three important qualities (\textit{readability}, \textit{informativeness}, and \textit{faithfulness}), we have seen that increasing the model parameters in LLMs inside of an LLM family can boost \textit{readability} and \textit{informativeness}, but it tends to reduce the \textit{faithfulness} of the D2T task.
Therefore, in scenarios where \textit{faithfulness} is crucial, such as in safety-critical applications like medical report generation~\citep{pauws2019making,hommes2019personalized,nishino2020reinforcement}, it is advisable to use a LLM from smaller model-sized LLM family in D2T applications.
However, in cases where \textit{readability} (fluency and coherence) and \textit{informativeness} are paramount, increasing parameters is a viable approach.

\section{Analyzing the Effect of Source-Reference Divergence}
\label{sec:effect_divergence}
As discussed in Subsection~\ref{subsec:divergence}, source-reference divergence is a common phenomenon in D2T tasks and is unavoidable in D2T datasets~\citep{dhingra2019handling,wiseman2017challenges,islam2023tackling}.
In this analysis, we examine two aspects: how LLMs perform in D2T tasks when source-reference divergence is present, and whether the number of parameters in LLMs affects their performance in the context of such source-reference divergence.
For this analysis, we consider three LLM families---T5, BLOOM, and Llama 2---along with six LLMs: \textsf{T5-base}, \textsf{T5-large}, \textsf{BLOOM-3B}, \textsf{BLOOM-7B}, \textsf{Llama2-7B}, and \textsf{Llama2-13B}.
Additionally, we incorporate all three types of D2T tasks, using one dataset for each type: E2E for MR-to-Text, WikiTableText for Table-to-Text, and DART for Graph-to-Text. 
We evaluate the performance of these models based on the three important qualities of D2T: \textit{readability} (measured by \textsc{Bleu}), \textit{informativeness} (measured by \textsc{MoverScore} and \textsc{Parent}), and \textit{faithfulness} (measured by \textsc{BartScore}).
To provide a clearer illustration of the impact of source-reference divergence, we categorize the test partitions of each dataset into three groups based on their average source-reference divergence: \texttt{low}, \texttt{medium}, and \texttt{high}.
The \texttt{low} group consists of instances with low source-reference divergences, while the \texttt{high} group comprises instances with high source-reference divergences.
Source-reference divergence ($\text{div} (s, r)$) is defined based on the differences between the source text ($s$) and the reference text ($r$) at the unigram level, as follows:

\begin{align*}
    \text{div} (s, r) = 1 - \frac{\big\vert\text{unigram}(s) \cap \text{unigram}(r)\big\vert}{\big\vert\text{unigram}(s) \cup \text{unigram}(r)\big\vert}
\end{align*}

Where $\vert x \vert$ denotes the cardinality and `$\text{unigram}(x)$' represents the set of all unigrams of the text $x$. 
We opt for unigram-based matching to calculate divergence rather than higher-order n-gram matching because the source data $S$ often consists of semi-structured non-textual data, where higher-order grams may overlook significant portions of information.

\begin{figure}[ht]
    \centering
    \resizebox{\textwidth}{!}{
    \includegraphics{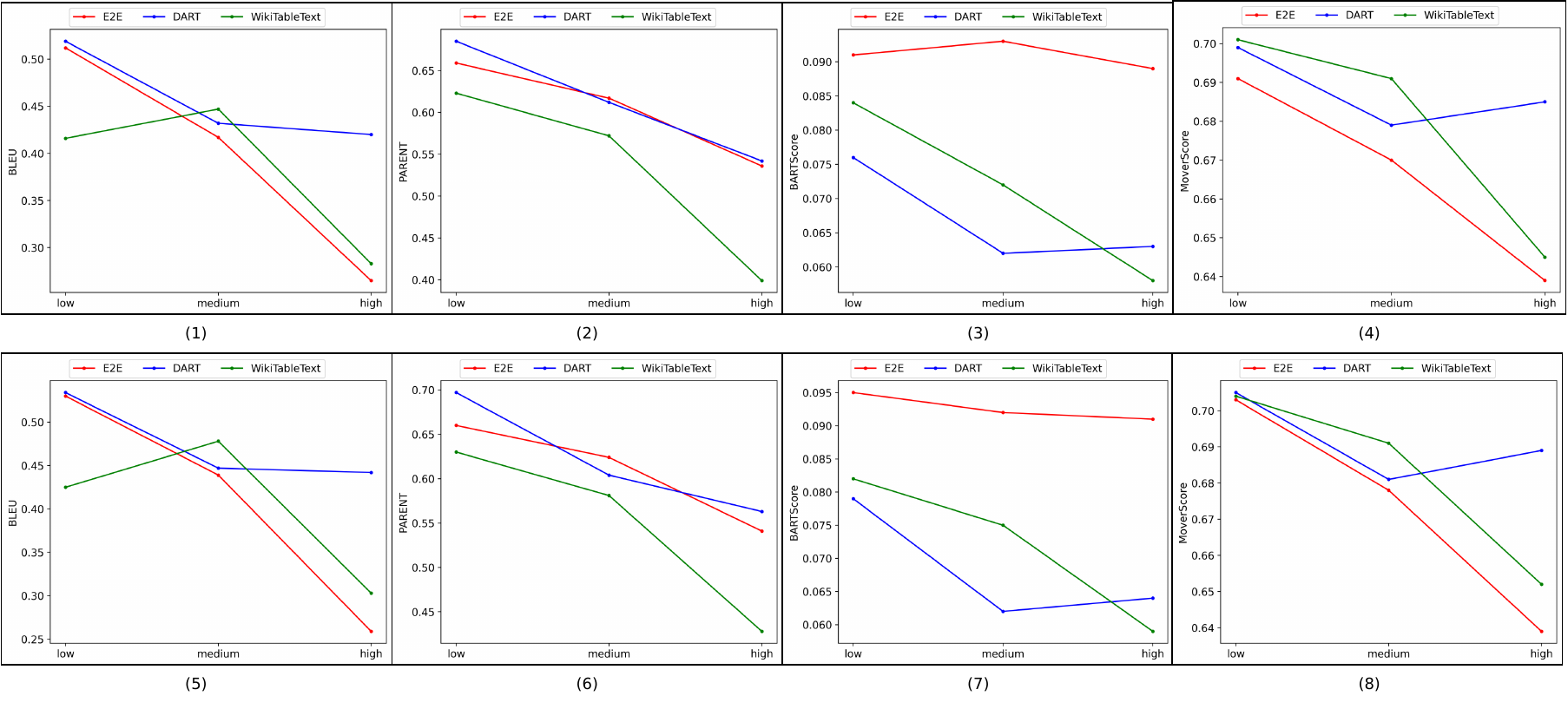}}
    \caption{Analysis of performance of T5 in context of source reference divergence based on three datasets (E2E, WikiTableText and DART) in terms of \textsc{Bleu}, \textsc{Parent}, \textsc{BartScore} and \textsc{MoverScore}. First row represents performance of T5-base, and second row represents performance of T5-large.}
    \label{fig:t5-divergence}
\end{figure}

\subsection{Influence of Source-Reference Divergence on LLM Performance for D2T}
\begin{figure}[ht]
    \centering
    \resizebox{\textwidth}{!}{
    \includegraphics{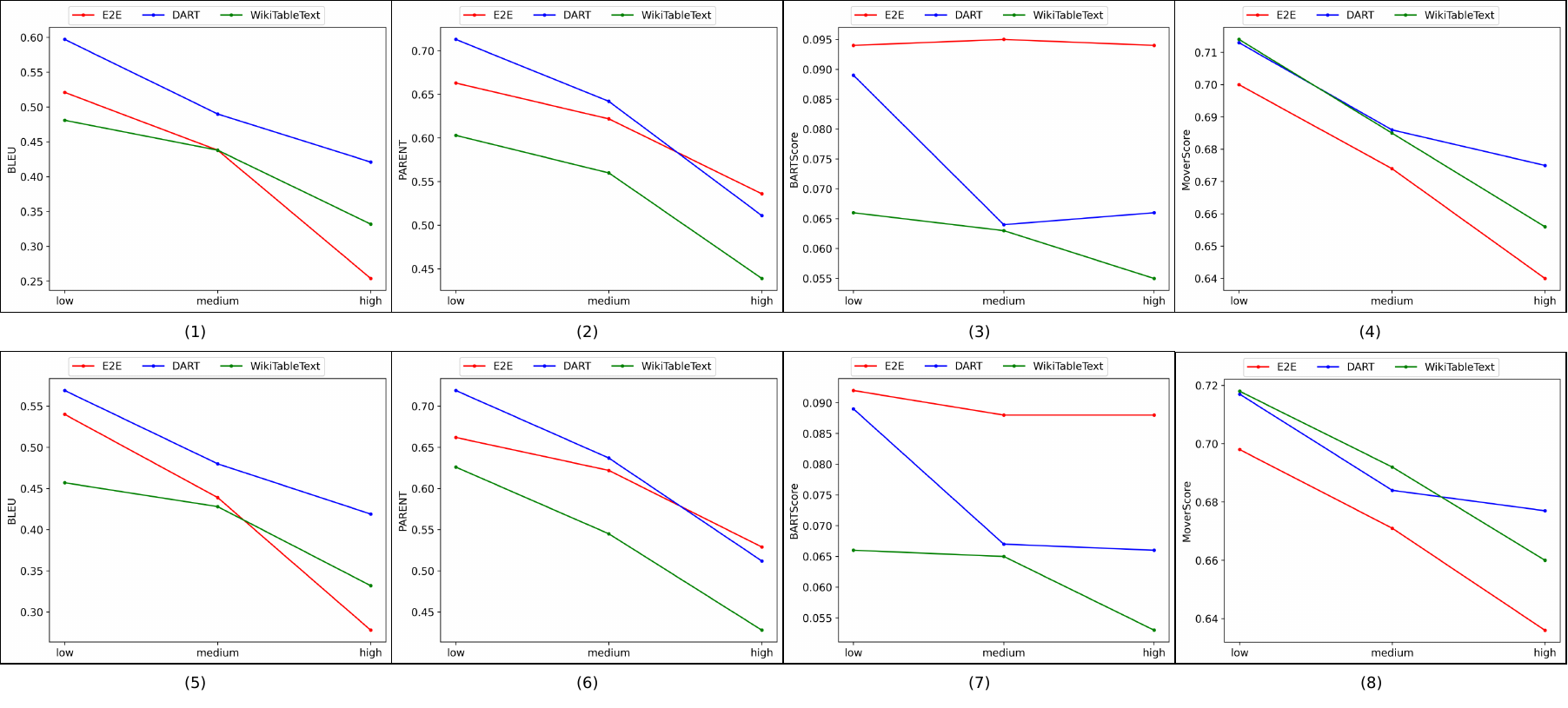}}
    \caption{Analysis of performance of Llama 2 in context of source reference divergence based on three datasets (E2E, WikiTableText and DART) in terms of \textsc{Bleu}, \textsc{Parent}, \textsc{BartScore} and \textsc{MoverScore}. First row represents performance of Llama2-7B, and second row represents performance of Llama2-13B.}
    \label{fig:llama2-divergence}
\end{figure}

In this part of analysis, we investigated the performance of LLMs under source-reference divergence.
Figures~\ref{fig:t5-divergence}, ~\ref{fig:t5-divergence}, and \ref{fig:llama2-divergence} illustrate the performance of T5, BLOOM, and Llama 2 across three D2T datasets in the context of source-reference divergence.
These figures provide insights to address the first part of the third question raised in Section~\ref{sec:aim_motivation}.

\begin{figure}[ht]
    \centering
    \resizebox{\textwidth}{!}{
    \includegraphics{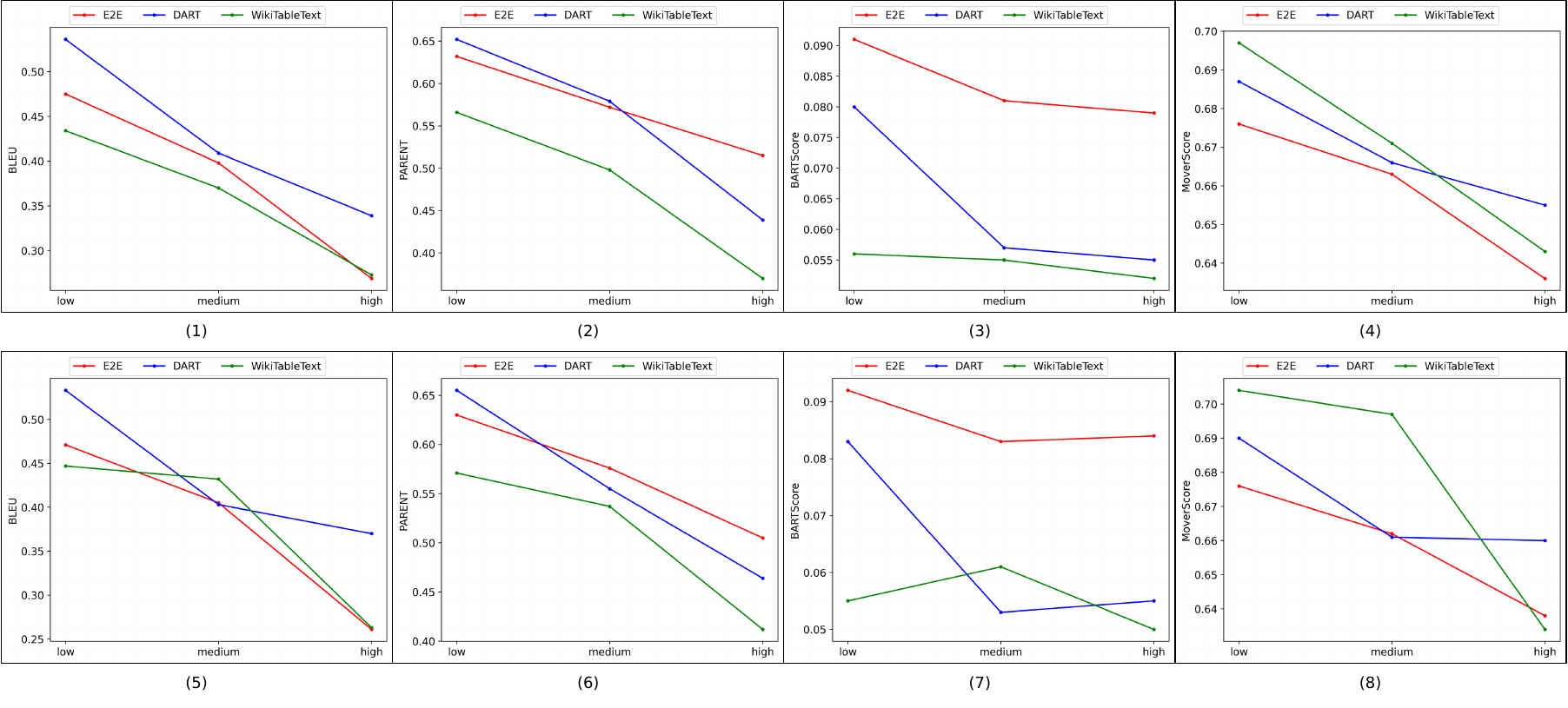}}
    \caption{Analysis of performance of BLOOM in context of source reference divergence based on three datasets (E2E, WikiTableText and DART) in terms of \textsc{Bleu}, \textsc{Parent}, \textsc{BartScore} and \textsc{MoverScore}. First row represents performance of BLOOM-3B, and second row represents performance of BLOOM-7B.}
    \label{fig:bloom-divergence}
\end{figure}

\begin{itemize}
    \item [] \textbf{Question:} \ul{Does the presence of source-reference divergence influence the performance of LLMs for D2T tasks?} \textcolor{gray}{If so, does increasing the model size of LLM aid in mitigating the effects of source-reference divergence?}
    \item [] \textbf{Answer:} The analysis of Figures~\ref{fig:t5-divergence}, ~\ref{fig:bloom-divergence}, and ~\ref{fig:llama2-divergence} reveals a significant influence of source-reference divergence on the performance of LLMs across all three LLM families. 
    Across the \texttt{low}, \texttt{medium}, and \texttt{high} divergence groups, distinct performance trends of LLMs emerge.
    In the \texttt{low} divergence group, characterized by minimal source-reference discrepancies, all LLM models consistently exhibit higher scores across the four evaluated metrics (\textsc{Bleu}, \textsc{Parent}, \textsc{MoverScore}, and \textsc{BartScore}). 
    This suggests that when the source-reference divergence is low, LLMs tend to perform better in terms of \textit{readability}, \textit{informativeness}, and \textit{faithfulness}.
    Conversely, in the \texttt{high} divergence group where source-reference discrepancies are more pronounced, all LLMs demonstrate notably lower performance across the metrics compared to the \texttt{low} and \texttt{medium} groups. 
    This observation underscores the significant impact of source-reference divergence on LLM performance, with higher levels of divergence leading to decreased performance across the evaluated qualities.
    In summary, these findings confirm that LLM performance is indeed influenced by source-reference divergence, with lower levels of divergence consistently associated with superior performance across the evaluated metrics.
    So, despite the claims of LLMs to possess greater generalization and knowledge, their performance in D2T tasks remains inadequate when source-reference divergence is present.
\end{itemize}

\subsection{Impact of model Size on LLM Performance for D2T in the Presence of Source-Reference Divergence}
In this section, we conduct an analysis to determine the influence of underlying parameter numbers (or simply size) on the performance of LLMs in the context of source-reference divergence.
To facilitate this analysis, we follow a similar methodology as in our previous analysis, segmenting the entire test set into three groups (\texttt{low}, \texttt{medium}, and \texttt{high}) based on source-reference divergence.
For comparison purposes, we juxtapose two LLMs in a single plot, specifically focusing on three comparisons: \textsf{Llama2-7B} vs \textsf{T5-base}, \textsf{Llama2-13B} vs \textsf{T5-base}, and \textsf{BLOOM-7B} vs \textsf{T5-base}. 
These comparisons enable us to juxtapose the performance of three larger LLMs (\textsf{BLOOM-7B}, \textsf{Llama2-7B}, and \textsf{Llama2-13B}) against that of a relatively smaller LLM, \textsf{T5-base}.
We analyze these comparisons across four metrics (\textsc{Bleu}, \textsc{Parent}, \textsc{MoverScore}, and \textsc{BartScore}), encompassing all three key qualities of D2T models over the three D2T datasets---E2E, WikiTableText, and DART.
Through these comparative analyses, we aim to address the final part of the third question raised in Section~\ref{sec:aim_motivation}, shedding light on the relationship between LLM parameter numbers and performance in the presence of source-reference divergence.

\begin{figure}[ht]
    \centering
    \resizebox{\textwidth}{!}{
    \includegraphics{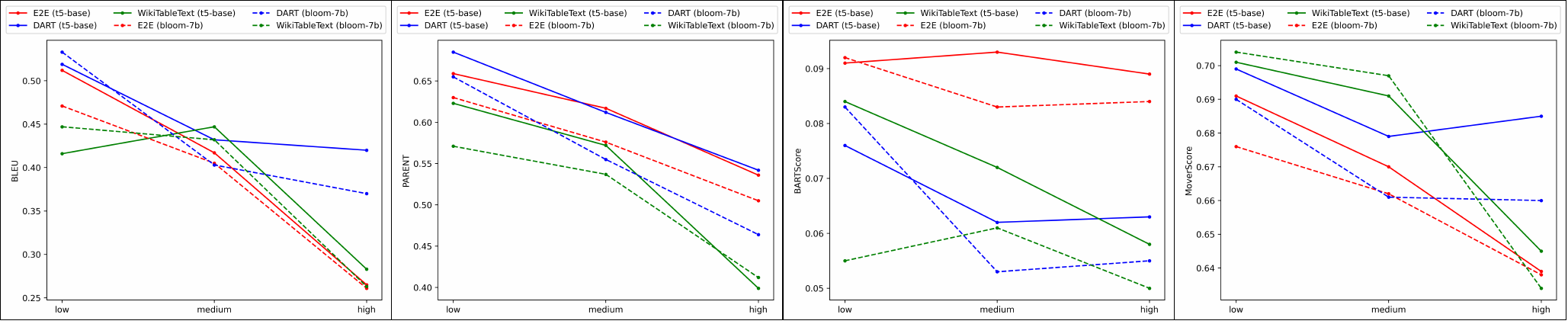}}
    \caption{Comparative performance analysis between T5-base and BLOOM-7B in context of source reference divergence based on three datasets (E2E, WikiTableText and DART) in terms of \textsc{Bleu}, \textsc{Parent}, \textsc{BartScore} and \textsc{MoverScore}.}
    \label{fig:t5_bloom7}
\end{figure}

\begin{itemize}
    \item [] \textbf{Question:} \textcolor{gray}{Does the presence of source-reference divergence influence the performance of LLMs for D2T tasks?} \ul{If so, does increasing the model size of LLM aid in mitigating the effects of source-reference divergence?}

    \begin{figure}[ht]
        \centering
        \resizebox{\textwidth}{!}{
        \includegraphics{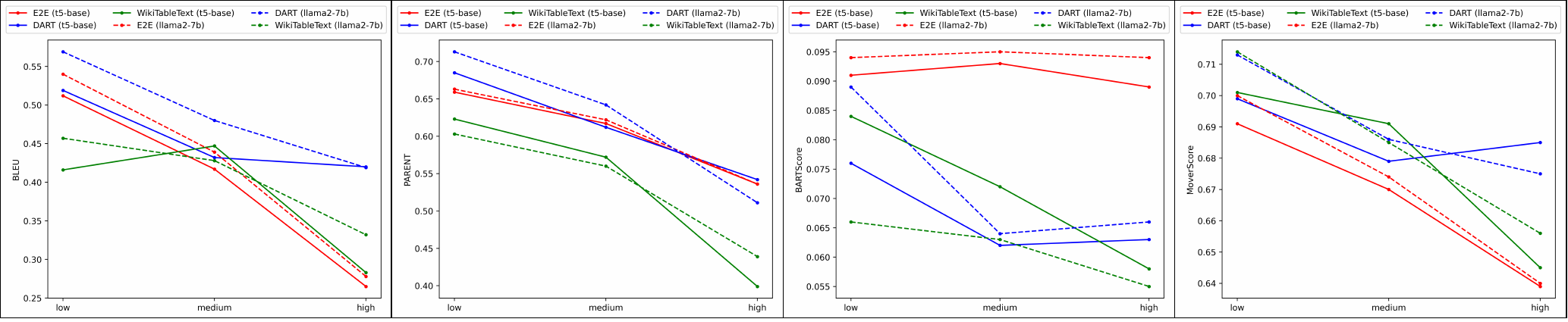}}
        \caption{Comparative performance analysis between T5-base and Llama2-7B in context of source reference divergence based on three datasets (E2E, WikiTableText and DART) in terms of \textsc{Bleu}, \textsc{Parent}, \textsc{BartScore} and \textsc{MoverScore}.}
        \label{fig:t5_llama7}
    \end{figure}

    \begin{figure}[ht]
        \centering
        \resizebox{\textwidth}{!}{
        \includegraphics{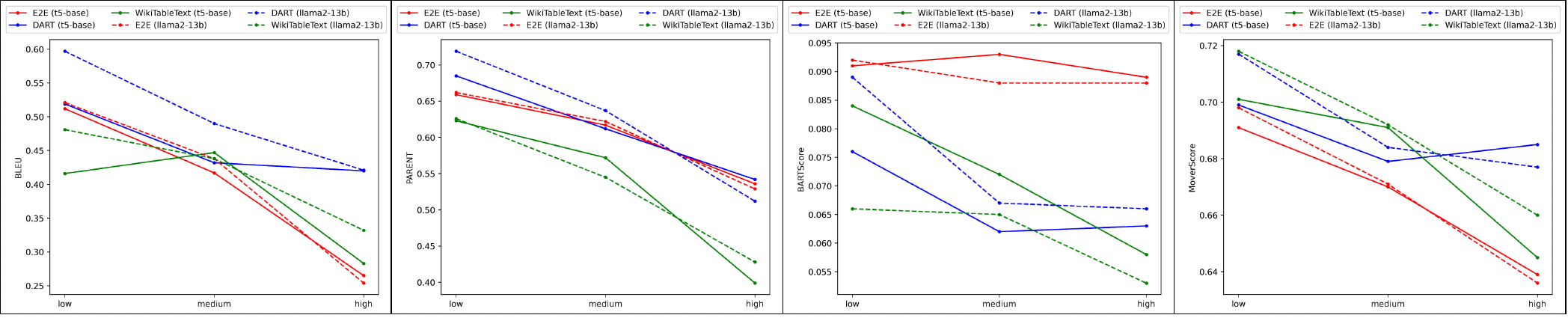}}
        \caption{Comparative performance analysis between T5-base and Llama2-13B in context of source reference divergence based on three datasets (E2E, WikiTableText and DART) in terms of \textsc{Bleu}, \textsc{Parent}, \textsc{BartScore} and \textsc{MoverScore}.}
        \label{fig:t5_llama13}
    \end{figure}
    
    \item [] \textbf{Answer:} From the comparison plots~\ref{fig:t5_bloom7}, ~\ref{fig:t5_llama7}, and ~\ref{fig:t5_llama13}, it becomes apparent that LLMs with larger parameters (\textsf{Llama2-13B}, \textsf{Llama2-7B}, and \textsf{BLOOM-7B}) tend to outperform low-parameter models like T5-base in D2T tasks when source-reference divergence is low.
    However, there are some exceptions, particularly in comparisons with the \textsf{BLOOM-7B} and \textsf{T5-base} models.
    As the source-reference divergence rate rises (i.e., in the \texttt{medium} group of source-reference divergence), \textsf{T5-base}, being a low-parameter model, begins to perform well and competes with the larger LLMs.
    Finally, when the source-reference divergence becomes high, T5-base often either outperforms the larger LLMs (\textsf{Llama2-13B}, \textsf{Llama2-7B}, and \textsf{BLOOM-7B}) in almost all metrics or significantly narrows the performance gap with them.
    Although there are some exceptional cases where \textsf{T5-base} cannot outperform LLMs, this is primarily observed with the WikiTableText dataset.
    Nevertheless, in the \texttt{medium} group of source-reference divergence in the WikiTableText dataset, the T5-base model often performs better than all three larger LLMs.
    These observations indicate that when source-reference divergence is high, it is advisable to use smaller LLMs like \textsf{T5-base} for better performance.
\end{itemize}

\begin{table}[]
    \centering
    \resizebox{\textwidth}{!}{
    \begin{tabular}{@{}ccccc@{}}
    \toprule
    source & reference & LLM & generated text & remarks \\ \midrule
    \multirow{4}{*}{\begin{tabular}[c]{@{}c@{}}name=The Vaults; eatType = pub; \\ food=French; priceRange=less than £20; \\ area=city centre; familyFriendly=yes\end{tabular}} & \multirow{4}{*}{\begin{tabular}[c]{@{}c@{}}Near Raja Cuisine, in the city centre \\ is a pub offering French food. The \\ Vaults is cheap and family-friendly.\end{tabular}} & T5-base & \begin{tabular}[c]{@{}c@{}}The Vaults is a family-friendly \\ \textcolor{red}{French pub} in the city centre \\ with a price range of less than £20.\end{tabular} & \multirow{4}{*}{\begin{tabular}[c]{@{}c@{}}All LLMs are successfully generated similar \\ texts, however T5-base transformed with \\ `food=French' into  \textcolor{red}{`French pub'}, mistakenly.\end{tabular}} \\ \cmidrule(lr){3-4}
     &  & T5-large & \begin{tabular}[c]{@{}c@{}}The Vaults is a family-friendly pub \\ in the city centre that \\ serves French food for less than £20.\end{tabular} &  \\ \cmidrule(lr){3-4}
     &  & Llama2-7B & \begin{tabular}[c]{@{}c@{}}The Vaults is a family-friendly\\  pub that serves French food. \\ It is located in the city centre \\ and has a price range of less than £20.\end{tabular} &  \\ \cmidrule(lr){3-4}
     &  & Llama2-13B & \begin{tabular}[c]{@{}c@{}}The Vaults is a family-friendly pub that\\  serves French food for less than £20. \\ It is located in the city centre.\end{tabular} &  \\ \midrule
    \multirow{4}{*}{\begin{tabular}[c]{@{}c@{}}name=The Waterman; eatType=restaurant; \\ priceRange=high; area=riverside; \\ familyFriendly=yes; near=Raja Indian Cuisine\end{tabular}} & \multirow{4}{*}{\begin{tabular}[c]{@{}c@{}}The Waterman is a restaurant near to \\ Raja Indian Cuisine that is \\ \textcolor{blue}{child-friendly}. It has a high price \\ range but is located in riverside.\end{tabular}} & T5-base & \begin{tabular}[c]{@{}c@{}}The Waterman is a high priced, \\ \textcolor{blue}{child friendly} restaurant in \\ the riverside area near Raja \\ Indian Cuisine.\end{tabular} & \multirow{4}{*}{\begin{tabular}[c]{@{}c@{}}In this example, there is a discrepancy between \\ the source data and the reference. The source \\ data mentions familyFriendly=yes, while the \\ reference uses the term \textcolor{blue}{`child-friendly'}. \\ Interestingly, all three LLMs, except Llama2-13B, \\ successfully identify this discrepancy. This case \\ highlights the possibility that larger LLMs may \\ overlook certain divergences in the data.\end{tabular}} \\ \cmidrule(lr){3-4}
     &  & T5-large & \begin{tabular}[c]{@{}c@{}}The Waterman is a \textcolor{blue}{child friendly} \\ restaurant in the riverside area \\ near Raja Indian Cuisine. \\ It has a high price range.\end{tabular} &  \\ \cmidrule(lr){3-4}
     &  & Llama2-7B & \begin{tabular}[c]{@{}c@{}}The Waterman is a high priced \\ restaurant in the riverside area \\ near Raja Indian Cuisine. \\ It is \textcolor{blue}{child friendly}.\end{tabular} &  \\ \cmidrule(lr){3-4}
     &  & Llama2-13B & \begin{tabular}[c]{@{}c@{}}The Waterman is a high priced family \\ friendly restaurant in the riverside area \\ near Raja Indian Cuisine.\end{tabular} &  \\ \bottomrule
    \end{tabular}}
    \caption{Case study on the E2E dataset featuring four LLMs (T5-base, T5-large, Llama2-7B, and Llama2-13B), illustrating errors made by smaller LLMs such as T5-base and instances where larger LLMs like Llama2-13B fail to identify source-reference divergence.}
    \label{tab:e2e_case_study}
\end{table}

\begin{table}[ht]
    \centering
    \resizebox{\textwidth}{!}{
    \begin{tabular}{@{}ccccc@{}}
    \toprule
    source & reference & LLM & generated text & remarks \\ \midrule
    \multirow{4}{*}{\begin{tabular}[c]{@{}c@{}}{[}{[}'subjtitle', 'history of the jews in malaysia'{]}, \\ {[}'subjsubtitle', 'history'{]},  {[}'year', '1881'{]}, \\ {[}'males', '14'{]}, {[}'females', '9'{]}, {[}'boys', '5'{]}, \\ {[}'girls', '4'{]}, {[}'total', '32'{]}{]}\end{tabular}} & \multirow{4}{*}{\begin{tabular}[c]{@{}c@{}}in 1988 , there were 32 the jews in \\ malaysia in which there were \\ 14 males , 9 females , 5 boys and \\ 4 girls .\end{tabular}} & T5-base & in 1881, there were 32 jews in malaysia. & \multirow{4}{*}{\begin{tabular}[c]{@{}c@{}}All LLMs are experiencing issues with \\ \textit{informativeness }as they struggle to generate \\ all the source content accurately. In this \\ example, larger LLMs show improved \\ \textit{informativeness}, with Llama2-13B achieving \\ comparatively better results.\end{tabular}} \\ \cmidrule(lr){3-4}
     &  & T5-large & \begin{tabular}[c]{@{}c@{}}in 1881, there were 14 males and \\ 9 females in malaysia.\end{tabular} &  \\ \cmidrule(lr){3-4}
     &  & Llama2-7B & \begin{tabular}[c]{@{}c@{}}there were 14 males and 9 females \\ in the jews in malaysia in 1881 .\end{tabular} &  \\ \cmidrule(lr){3-4}
     &  & Llama2-13B & \begin{tabular}[c]{@{}c@{}}there were 14 males , 9 females , \\ 5 boys and 4 girls in the jews \\ in malaysia in 1881 .\end{tabular} &  \\ \midrule
    \multirow{4}{*}{\begin{tabular}[c]{@{}c@{}}{[}{[}'subjtitle', 'bath rugby'{]}, {[}'subjsubtitle', 'current squad'{]}, \\ {[}'player', 'jack wilson'{]}, {[}'position', 'wing'{]}, \\ {[}'union', 'new zealand'{]}{]}\end{tabular}} & \multirow{4}{*}{\begin{tabular}[c]{@{}c@{}}in bath rugby , wing jack wilson \\ was a member of new zealand union .\end{tabular}} & T5-base & jack wilson was a \textcolor{blue}{wing} in new zealand. & \multirow{4}{*}{\begin{tabular}[c]{@{}c@{}}While T5-base and T5-large provide \textcolor{blue}{faithful} \\ \textcolor{blue}{information}, Llama2-7B and Llama2-13B \\ outperform them in terms of \textit{readability}.\end{tabular}} \\ \cmidrule(lr){3-4}
     &  & T5-large & jack wilson was from \textcolor{blue}{union} new zealand. &  \\ \cmidrule(lr){3-4}
     &  & Llama2-7B & jack wilson was from new zealand . &  \\ \cmidrule(lr){3-4}
     &  & Llama2-13B & jack wilson was from new zealand . &  \\ \bottomrule
    \end{tabular}}
    \caption{Case study on the WikiTableText dataset featuring four LLMs (T5-base, T5-large, Llama2-7B, and Llama2-13B), demonstrating how increasing LLM size enhances \textit{informativeness}, while smaller LLMs like T5-base and T5-large generate faithful but less readable content compared to larger LLMs.}
    \label{tab:wikitabletext_case_study}
\end{table}

\section{Case Studies}
\label{sec:case_studies}
We present two case studies (Table~\ref{tab:e2e_case_study} and~\ref{tab:wikitabletext_case_study}) conducted on the E2E and WikiTableText datasets, each showcasing two samples and their corresponding generations from four LLMs: \textsf{T5-base}, \textsf{T5-large}, \textsf{Llama2-7B}, and \textsf{Llama2-13B}.
In the first case (Table~\ref{tab:e2e_case_study}) study using the E2E dataset, one example illustrates how smaller LLMs like \textsf{T5-base} may fail to accurately convey information. 
Another example highlights a source-reference divergence phenomenon, where larger LLMs like \textsf{Llama2-13B} fail to capture the divergence, while other LLMs successfully handle it.
Therefore, Table~\ref{tab:e2e_case_study} demonstrates that smaller-sized LLMs may sacrifice \textit{informativeness}, while larger LLMs frequently struggle to maintain source-reference divergence.
The second case study (Table~\ref{tab:wikitabletext_case_study}) on the WikiTableText dataset demonstrates how increasing LLM size enhances \textit{informativeness} as a D2T model. 
Additionally, it reveals that smaller LLMs such as \textsf{T5-base} and \textsf{T5-large} tend to prioritize \textit{faithfulness}, while larger LLMs prioritize \textit{readability}.
Table~\ref{tab:wikitabletext_case_study} reveals that smaller-sized LLMs prioritize \textit{faithfulness}, rendering them more suitable for applications requiring safety-critical considerations.

\section{Conclusion}
\label{sec:conclusion}
This paper investigates the impact of model size (number of model parameters) on the performance of fine-tuned LLMs in D2T generation tasks, focusing on three key qualities: \textit{readability}, \textit{informativeness}, and \textit{faithfulness}. 
Despite its significant importance, this investigation has been entirely overlooked in the existing literature on LLMs for D2T applications.
We conducted a comprehensive assessment involving twelve LLMs drawn from five popular LLM families: T5, BART, OPT, BLOOM, and Llama 2. 
Our analysis encompasses three primary D2T task categories: graph-to-text, table-to-text, and MR-to-text.
Six widely recognized evaluation metrics to ensure a thorough examination of D2T models: \textsc{BLEU}, \textsc{METEOR}, \textsc{BERTScore}, \textsc{MoverScore}, \textsc{Parent}, and \textsc{BARTScore}.
Given the inherent risks associated with human evaluation~\citep{freitag2021experts}, such as inter-annotator knowledge disparities and cognitive biases, we have opted not to include human evaluation results in our study to ensure unbiased comparisons. 
Nonetheless, two case studies are presented to enhance clarity in our analyses.
Moreover, we examine a crucial aspect of D2T tasks: source-reference divergence, which presents challenges in generating a reference from the source data.
Our primary findings can be summarized into three main points. 
Firstly, we find that increasing the parameters of large language models (LLMs) within a given LLM family generally improves \textit{readability} and \textit{informativeness} in data-to-text (D2T) tasks.
However, this often comes at the expense of \textit{faithfulness} in such tasks, suggesting a preference for smaller LLMs in safety-critical D2T domains.
Secondly, we demonstrate that larger-sized LLM families do not consistently outperform smaller ones in D2T task performance, challenging the notion that bigger models always yield better results.
Lastly, in the context of source-reference divergence, we meticulously observe that the performance of all LLMs is negatively affected.
However, smaller LLMs exhibit greater resilience compared to their larger counterparts in dealing with source-reference divergence.
We firmly believe that our research offers a profound understanding of the nuanced interplay between model size and LLM performance in D2T tasks, furnishing valuable insights to optimize the utilization of LLMs across diverse D2T applications.

\section*{Acknowledgment}
This research is partially supported by the Indo-French Centre for the Promotion of Advanced Research (IFCPAR/CEFIPRA) through CSRP Project No. 6702-2.

\bibliography{references/final_base} 
\bibliographystyle{plainnat}
\end{document}